\documentclass{opt2020} 

\makeatletter
\def\footnotecomment{\gdef\@thefnmark{}\@footnotetext}
\makeatother

\def\R{\mathbb{R}}

\def\d{\mathrm{d}}
\DeclareMathOperator{\Div}{div}
\def\calL{\mathcal{L}}
\def\calA{\mathcal{A}}
\newcommand{\ps}[1]{\langle #1 \rangle}
\newcommand{\psbig}[1]{\big\langle #1 \big\rangle}
\newcommand{\intMean}[1]{\langle #1 \rangle}

\newcommand{\norm}[1]{\| #1 \|}

\def\manifold{\Sigma}
\def\frakg{\mathfrak{g}}
\def\calM{\mathcal{M}}
\def\loss{L_X}

\def\Ld{L^2}

\def\Ldznu{L^2_0(\nu_\manifold)}
\def\fcnC{\mathcal{C}}
\def\E{\mathbb{E}}
\def\d{{\rm d}}

\title[Constraint-Based Regularization of Neural Networks]{Constraint-Based Regularization of Neural Networks}

\optauthor{\Name{Benedict Leimkuhler} \Email{b.leimkuhler@ed.ac.uk}\\
  \Name{Timoth{\'e}e Pouchon} \Email{timothee.pouchon@ed.ac.uk}\\
    \Name{Tiffany Vlaar*} \Email{Tiffany.Vlaar@ed.ac.uk}\\
      \Name{Amos Storkey} \Email{a.storkey@ed.ac.uk}\\
  \addr University of Edinburgh, UK}

\begin{document}

\maketitle

\begin{abstract}%
We propose a method for efficiently incorporating constraints into a stochastic gradient Langevin framework for the training of deep neural networks.  Constraints allow direct control of the parameter space of the model. Appropriately designed, they reduce the vanishing/exploding gradient problem, control weight magnitudes and stabilize deep neural networks and thus improve the robustness of training algorithms and generalization capabilities of the trained neural network.  We present  examples of constrained training methods  motivated by orthogonality preservation for weight matrices and  explicit weight normalizations. We describe the methods in the overdamped formulation of Langevin dynamics  and the underdamped form, in which momenta help to improve sampling efficiency. Our methods see performance improvements on image classification tasks.
\end{abstract}




\footnotecomment{\hspace*{-0.45cm} See \emph{``Better Training using Weight-Constrained Stochastic Dynamics"} (ICML 2021) for a later iteration of this work.}

\section{Introduction}
In this paper we explore stochastic training methods based on Langevin dynamics combined with algebraic constraints. Our general framework allows for incorporating constraints into standard training schemes and sampling methods for neural networks (NNs). Constraints provide direct control of the parameter space of a model and hence afford a means to improve its generalization performance. Current approaches to enhance the generalization performance of overparameterized NNs consist of both explicit and implicit regularization techniques \cite{neyshabur2015}. Examples of the former are L1 \cite{LASSO1, LASSO2} and L2 \cite{ridge} regularization, which modify the loss by adding a parameter norm penalty term. Batch normalization (BatchNorm) \cite{BatchNorm} is a technique that causes an implicit regularization effect. BatchNorm can be viewed as tantamount to a constraint imposed on the network's parameters during training. Although BatchNorm is widely used, explanations for the method's success remain elusive; claims that it would reduce internal covariance shift \cite{BatchNorm} or smooth the loss landscape \cite{BNhelpopt} have been disputed \cite{BNhelpopt,PyHessian}.  The reliance on increasingly complex strategies does little to enhance the explainability of NNs, so robust simplification of all aspects of  training is desirable.

In this paper we highlight the potential of constrained stochastic differential equation (SDE) based algorithms to provide a simpler overall framework for NN training. An example of a constraint that can be easily introduced using our general framework is orthogonality of the weight matrix. We provide a detailed algorithm for this purpose in a Langevin dynamics-based setting.

In NN training one aims to minimize the loss $L_X(\theta)$ for parameters $\theta\in \mathbb{R}^{|n|}$ and data $X$.  A popular training scheme is stochastic gradient descent (SGD).
SGD may be improved by incorporating momenta and additive noise, or more generally by embedding the loss gradient in a Langevin dynamics (LD) framework \cite{Cheng2017}.   Using low temperatures \cite{coldposterior, LMV}, sampling methods have been found to enhance exploration and speed the approach to `good' minima, which enhance their generalization to nearby data sets.  Ergodic properties of the idealized SDEs associated with gradient schemes may help these methods to ensure robust exploration of a useful range of parameters.

Constraints can be seen as limiting cases of penalty-based regularization which replaces minimization of the loss $L_X(\theta)$ by that of the augmented loss 
$L^c_X(\theta) = L_X(\theta) + \frac{1}{\varepsilon^2}g(\theta)^2$,
where $g(\cdot)$ is a suitable smooth function of the parameters. In the limit $\varepsilon\rightarrow 0$, these penalty terms introduce an undesirable stiffness and consequent stability restriction in gradient-based training.  It is therefore natural to relate the above system to a constrained optimization task subject to $g(\theta)=0$.  

\vspace*{-0.1cm}
\section{Neural networks with constraints} 	\label{sec:constraints}
\vspace*{-0.2cm}

We suggest to use constraints when training NNs.  Imposing good priors on NNs is known to improve performance, e.g. CNNs suit image datasets better than overparameterized fully connected NNs, despite being a subset of the latter \cite{haystack}. Using constraints also arises naturally in the control of vanishing/exploding gradients. Constraints can be used to control the magnitudes of individual weights and/or to limit the growth of gradients in deep NNs. 
We present various approaches below.

A $L$-layer NN has parameters $\theta\in\R^{|n|}$, with a weight matrix $W^{\ell} \in\R^{d^\ell\times d^{\ell-1}}$ and bias vector $b^\ell \in\R^{d^{\ell}}$ for each layer $\ell$.  
To allow for inequality constraints, we define slack variables vector $\xi\in\mathbb{R}^{n^\xi}$ 
and consider as variable $q = (\theta,\xi)\in\mathbb{R}^{d}$, where $d=|n|+n^\xi$.
The constraint manifold is 
\begin{equation}	\label{eq:constraintManifold}
\manifold = \{ q\in\R^d \mid g(q)=0  \}, \ \ g:\R^d\to\R^m.
\end{equation}
We partition $\theta = (\theta^u,\theta^c)$ into unconstrained $\theta^u\in\mathbb{R}^{n^u}$ and constrained $\theta^c\in\mathbb{R}^{n^c}$ parameters.
\par{\bf Circle constraints:}
In a {\em circle constraint}, we restrict each parameter in $\theta^c$ as 
$|\theta^c_i|\leq r_i$, where $r_i>0$ is given.
We thus introduce $m = n^c=n^\xi$ slack variables $\xi_i$
and define
\begin{equation}	\label{eq:circleConstraint}
g_i(q) = |\theta^c_i|^2 + |\xi_i|^2 - r_i^2 \qquad 1\leq i\leq m.
\end{equation}
Note that if $q\in\manifold$ then the parameters in $\theta^c$ are bounded as desired.

\par{\bf Sphere constraints:} In a similar way, we could opt to restrict the sums of squares of weights associated to the input channels of any node. 
This constraint is analogous to max-norm \cite{maxnorm,dropout2} as used in ad hoc regularization procedures. In our context, introducing such constraints would yield distinctive training methods, although we omit discussion of these here due to space limitations.

\par{\bf Orthogonality constraints:}  
Orthogonal weight matrices can mitigate the vanishing/exploding gradient problem in RNNs \cite{PascanuRNNs,RNNsLongTermDependencies,UnitaryEvolutionRNNs}, and are developing a growing following in the CNN literature
\cite{OrthReg_CNNs,Rodriguez2017, CNN_stiefelmanifold}. Orthogonal initialization is linked to achieving dynamical isometry \cite{Saxe2013, Pennington2017, Pennington2018}, which can accelerate training. \citet{10000CNN} were able to train 10,000 layer vanilla CNNs, 
without learning rate decay, BatchNorm or residual connections, by using initial orthogonal convolution kernels. Methods for enforcing orthogonality during training include the use of `soft' constraints which add a restraint term to the loss \cite{BeyondGoodInit,OrthReg_CNNs, NeuralPhotoEditing} and hard constraints based on optimization over Stiefel manifolds \cite{CNN_stiefelmanifold, OrthDNN}. The latter requires repeated singular value decomposition of high-dimensional matrices during training, which is costly. Here we propose a straightforward algorithm to incorporate orthonormality constraints for rectangular matrices within our NN training framework, with manageable additional cost. We make no empirical claims over other manifold optimization methods, but rather provide a framework for network optimization that is theoretically sound, flexible enough to incorporate new constraints, and demonstrates good properties relative to standard SGD training. 
We set $\theta^u = b^\ell$, 
and define the \textit{orthogonality constraint} for layer $\ell$ (which has $n^{\ell}$ parameters) as 
\begin{equation}	\label{eq:orthogonalConstraint}
g(q) =
\left\{
\begin{array}{ll}
\big(W^\ell\big)^T W^\ell - I_{n^{\ell-1}} 	&\text{if }n^{\ell-1}\leq n^{\ell}, \\
W^\ell \big(W^\ell\big)^T  - I_{n^{\ell}} 	&\text{otherwise}.
\end{array}
\right.
\end{equation}

\vspace*{-0.4cm}
\section{Constrained SDEs and their discretization}	\label{sec:constrainedSDEs}	
\vspace*{-.2cm}
We now describe SDE-based methods for constrained NN training.   An alternative to our approach is constrained Hamiltonian Monte Carlo (HMC) methods \cite{ZHG2018,GS2017,LSZ2020}.
Although HMC schemes have nil sampling bias if fully converged, their acceptance rates depend on stepsize and system size \cite{Beskos2013,BRSS2017}. SDE-based methods are often preferred in high-dimensional sampling calculations as they offer greater overall efficiency for a fixed computational budget. In this section we discuss properties of constrained Langevin Dynamics. 
For further discussion on (unconstrained) LD see \cite{Pav14}. 
 LD discretizations are studied in \cite[Chap. 3]{LSR10}, \cite{FaL09} (overdamped) and \cite{LRS12,LeM16} (underdamped). 

\vspace*{-0.1cm}
\paragraph{Constrained Langevin: ergodicity and central limit theorem.}
The NN loss function naturally extends to the variable $q=(\theta,\xi)\in\R^d$ 
as $V(q) = \loss(\theta)$ (note that in particular $\nabla_\xi V = 0$). 
The first continuous training method we consider is the constrained overdamped Langevin system
\begin{equation}
	\label{eq:constrainedOverdampedLangevin}
	\d q_t = -\nabla V(q_t)\,\d t +  \sqrt{2\tau}\,\d \mathcal{W}_t  - \nabla_q g(q_t)\,\d \lambda_t, 
	\qquad 0 = g(q_t),
\end{equation}
where $\mathcal{W}$ is a $d$-dimensional Wiener process, $\tau\geq 0$ is the temperature hyperparameter,
and $\lambda_t$ is an $\R^m$-valued vector of Lagrange multipliers.
Provided the initial configuration $q_0$ satisfies the constraint, any trajectory $q_t$ of \eqref{eq:constrainedOverdampedLangevin} remains on the constraint manifold $\manifold$ defined in \eqref{eq:constraintManifold}.
When $\beta^{-1}=\tau>0$, \eqref{eq:constrainedOverdampedLangevin} is equivalent to an underlying  ergodic (unconstrained) SDE (see \cite[Chap. 3]{LSR10} and Appx. \ref{sec:proofUnderlyingSDE}) with unique invariant measure 
$\d\nu_\manifold = Z^{-1} e^{-\beta V(q)} \, \d\sigma_\manifold,
\ 
Z = \int_\manifold e^{-\beta V(q)} \, \d\sigma_\manifold,$
where $\sigma_\manifold$ is the surface measure on $\manifold$.
Ergodicity ensures that averages of observables with respect to $\nu_\manifold$ can be approximated by time averages of trajectories of \eqref{eq:constrainedOverdampedLangevin}. 
To ensure the practical use of \eqref{eq:constrainedOverdampedLangevin} as a training method, we need the convergence to occur in a reasonable time. 
Thanks to the reversibility of the underlying SDE (see Appx. \ref{sec:proofUnderlyingSDE}), exponential convergence to equilibrium occurs as a consequence of a Poincar\'e inequality for $\nu_\manifold$, which holds provided the curvature of the manifold is well behaved (see Appx. \ref{sec:proofPoincare} and \ref{sec:proofCorollaries}). 
Poincar\'e inequalities on manifolds and their use in the analysis of diffusion processes are presented in \cite[Chap. 4]{BGL13}.

Introducing momenta $p$ leads to constrained underdamped LD, 
the 2nd order counterpart of \eqref{eq:constrainedOverdampedLangevin}
	\begin{eqnarray}
	&&
	\begin{aligned}
	\d q_t &= p_t \,\d t,\ \ \ \d p_t = \big(-\nabla_q V(q_t) - \gamma p_t\big) \,\d t +  \sqrt{2\gamma\tau}\,\d \mathcal{W}_t  - \nabla_q g(q_t) \,\d \lambda_t, \ 0 = g(q_t),
	\end{aligned}
	\label{eq:constrainedUnderdampedLangevin_sde}
	\end{eqnarray}
where $\gamma$ is the friction hyperparameter. 
The constraint 
induces a cotangency condition:
$p\in T_q^*\manifold$, where $T_q^*\manifold = \{ p\in\R^d \mid \nabla^T g(q) p = 0\}$ is the cotangent space of the manifold $\manifold$. 
The corresponding phase space is the cotangent bundle 
$T^*\manifold = \{ (q,p) \mid q\in\manifold, p\in T_q^*\manifold\}$.
Given an initial pair $(q,p) \in T^*\manifold$, any trajectory $(q_t,p_t)$ of \eqref{eq:constrainedUnderdampedLangevin_sde}  stays on $T^*\manifold$ for all time. In case $\tau>0$,
\eqref{eq:constrainedUnderdampedLangevin_sde} is equivalent to an underlying ergodic SDE, whose invariant measure is 
$\d\mu = e^{-\beta H(q,p)} \d\sigma_{T^*\manifold}$, with Hamiltonian $H(q,p) = V(q) + \frac{1}{2} p^Tp$ and
$\sigma_{T^*\manifold}$ the Liouville measure of the cotangent bundle \cite{LRS12}.
Exponential convergence also holds here, but the proof is more technical (e.g. based on hypocoercivity \cite{Vil09, LelS16}).
\vspace*{-0.2cm}
\paragraph{Discretization of constrained Langevin dynamics.}
The simplest iteration scheme $q_n\in\manifold\mapsto q_{n+1}\in\manifold$ for constrained overdamped Langevin dynamics \eqref{eq:constrainedOverdampedLangevin} consists of an Euler--Maruyama step followed by projection onto  $\manifold$.
The best choice for the projection is constraint-specific. 
For circle constraints we suggest orthogonal projection, which is both explicit and robust (see Appx. \ref{sec:discr_OLcircle}).   
For orthogonality constraints, we derive an efficient quasi-Newton scheme (Appx. \ref{sec:OL_orthogonalConstraint}).
The latter leads to the following training method
(written here for $Q = W^\ell$ if $n^\ell\leq n^{\ell-1}$ and $Q = (W^\ell)^T$ otherwise, $s=\min\{n^\ell,n^{\ell-1}\}$):
one training iteration $Q_n\in\manifold\mapsto Q_{n+1}\in\manifold$ is given by
\begin{equation}	
\text{for $k = 0$ to $K-1$:} \qquad Q^{(k+1)} = Q^{(k)} - \tfrac{1}{2}Q_n\big( (Q^{(k)})^T Q^{(k)} - I_s \big).
\end{equation}
We initialize
$Q^{(0)} = Q_n - h\nabla_Q V(Q) + \sqrt{2\tau h} R_n$, with stepsize $h$ and independent standard random normal matrix $R_n$ of the same size as $Q$. 
After $K$ quasi-Newton iterations we set $Q_{n+1} \equiv Q^{(K)}$.
For the constrained underdamped Langevin system \eqref{eq:constrainedUnderdampedLangevin_sde}, the ABO splitting strategy from \cite{LeM16} gives 
\begin{eqnarray}
{\text{A:}} & & 
\d q_t = p_t \,\d t,\quad
\d p_t = - \nabla_qg(q_t)\,\d \lambda_t,\quad
0=g(q_t),\ \  0 = \nabla_qg(q_t)p_t, \label{eq:Acomponent}\\
{\text{B:}} & & 
\d q_t =0, \ 
\d p_t = -\nabla_q V(q_t)\,\d t -\nabla_q g(q_t)\,\d \mu_t,\quad
0=g(q_t),\ \  0 = \nabla_qg(q_t)p_t, \label{eq:Bcomponent}\\
{\text{O:}} & &
\d q_t =0, \  \d p_t = -\gamma p_t\,\d t + \sqrt{2\gamma \tau}\,\d \mathcal{W}_t -\nabla_q g(q_t)\,\d \nu_t, \ 
0=g(q_t), \ 0 = \nabla_qg(q_t)p_t\label{eq:Ocomponent}
\end{eqnarray}
We use an OBA sequence, which in the case $\tau=0$ and by re-scaling the momentum and step size variables, 
is equivalent to the standard PyTorch form of SGD with momentum \cite{Pytorch, LMV}. 
The B and O components can be solved exactly (in law) while the A component can be approximated using a standard scheme for constrained ODEs (e.g. SHAKE or RATTLE \cite[Chap. 7]{LeR04}). Importantly, the A component does not involve the evaluation of the gradient.
For circle constraints the A step can be solved explicitly (see Appx. \ref{sec:discr_ULcircle}).
For orthogonality constraints 
(Appx. \ref{sec:UL_orthogonalConstraint}):
for $Q\in\manifold$, the projection onto the cotangent space $T^*_Q\manifold$ is defined as 
$\Pi_Q : \R^{r\times s} \to \R^{r\times s}, \quad
\bar P \mapsto \Pi_Q\bar P = \bar P - \frac{1}{2} Q (\bar P^TQ + Q^T\bar P).$
Then the ABO steps $(Q_n,P_n)\in T^*\manifold \mapsto (Q_{n+1},P_{n+1})\in T^*\manifold$
are
\begin{equation}	\label{eq:UL_OG_main}
\begin{aligned}
&
\text{(\text{A})} \quad 
\left\{
\begin{aligned}
&Q^{(0)} = Q_n  + hP_{n}, \ \text{for $k = 0$:$K-1$:} ~\  Q^{(k+1)} = Q^{(k)} - \tfrac{1}{2}Q_n\big( (Q^{(k)})^T Q^{(k)} - I_s \big),\\ 
&Q_{n+1} = Q^{(K)},\ \bar P_{n+1} = P_n + \tfrac{1}{h}\big(Q_{n+1}-Q^{(0)}\big), 
\quad P_{n+1} = \Pi_{Q_{n+1}} \bar P_{n+1},
\end{aligned}
\right.
\\&
\text{(B)} \quad \  
\left\{
\begin{aligned}
&Q_{n+1} =Q_{n},
\qquad \bar P_{n+1} = P_n - h\nabla_Q V(Q_n),\quad
\quad P_{n+1} = \Pi_{Q_n}\bar{P}_{n+1},
\end{aligned}
\right.
\\&
\text{(\text{O})} \quad 
\left\{
\begin{aligned}
&Q_{n+1} =Q_{n},
\qquad \bar P_{n+1} = e^{-\gamma h} P_{n} + \sqrt{\tau(1-e^{-2\gamma h})}R_n,
\quad P_{n+1} = \Pi_{Q_n} \bar P_{n+1},
\end{aligned}
\right.
\end{aligned}
\end{equation}

\vspace*{-0.3cm}
\section{Numerical Experiments}
\vspace*{-0.2cm}
The use of constraints can enhance generalization performance. 
We support this claim by comparing the performance of NN architectures trained using our constrained methods vs. unconstrained SGD. We set $\tau=0$ and use equivalent learning rates to present a fair comparison. We denote our circle and orthogonal Constrained \textit{overdamped} Langevin Algorithms as c-CoLA-\textit{od} and o-CoLA-\textit{od}, respectively. We compare \textit{underdamped} variants (CoLA-\textit{ud}) with SGD with momentum (SGD-m). 

\begin{figure}[h]
    \centering
    \includegraphics[scale=0.4]{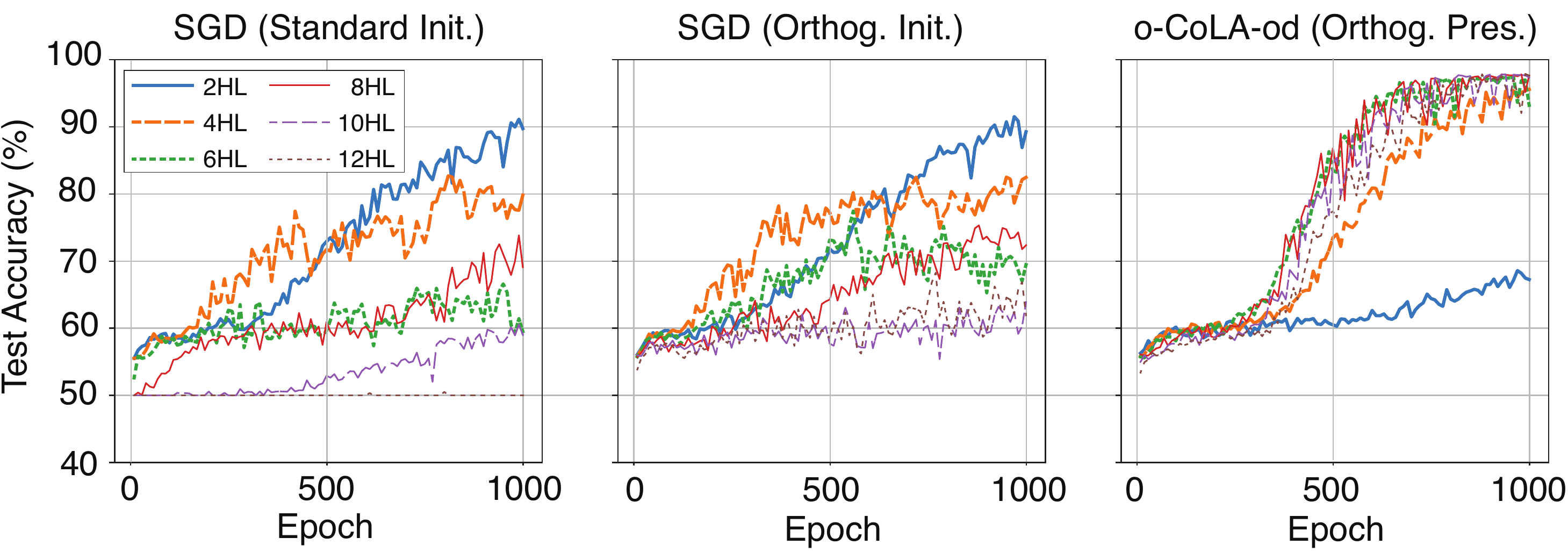}
    \caption{Test acc. 
    of MLPs with $p$-number of 100-node hidden layers (HL), ReLU activation. 
    The MLPs are trained on a 4-turn spiral dataset (Fig. C\ref{Spiraldata}) 
    using SGD with standard initialization (left), SGD with orthogonal initialization (middle) and o-CoLA-od with $ \tau = 0$ (right). For o-CoLA-od we constrain weights in all layers, apart from  input and output layers. Stepsize $h = 0.1$ for all methods. 
    Results are averaged over 10 runs. o-CoLA-od significantly outperforms unconstrained SGD for MLPs with more than 3 hidden layers.}\label{MLP_OCSGD}
\end{figure}

\begin{figure}[h]
    \centering
    \includegraphics[scale=0.45]{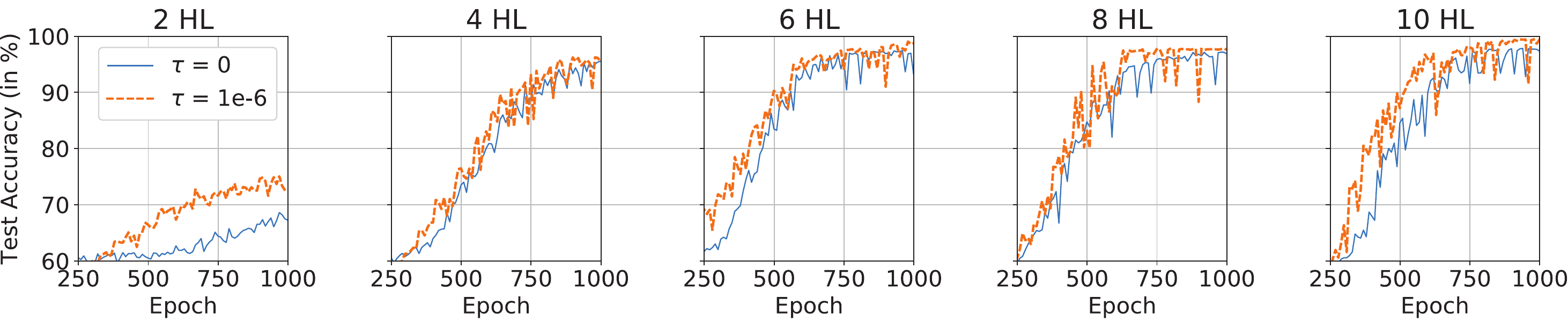}
    \caption{The effect of temperature for the same set-up as for Fig. \ref{MLP_OCSGD}. MLPs with varying numbers of hidden layers (HL) were trained using o-CoLA-od with $h = 0.1$ and either $\tau = 0$ (blue line) or $\tau$ = 1e-6 (orange line). Results are averaged over 5 runs. The use of temperature is shown to speed up training and often slightly increases the obtained test accuracies.}\label{Spiral_withT} \vspace*{-0.5cm}
\end{figure}

\paragraph{Orthogonality Constraints}
We compare SGD with orthogonality-preserving overdamped Langevin (Fig. \ref{MLP_OCSGD}). The goal is to train a MLP with $p$  hidden layers on a 
tightly wound spiral binary classification problem (Fig. C\ref{Spiraldata}). 
For SGD we show results for both i) standard PyTorch initialization and ii) orthogonal initialization. 
 A clear advantage imposing orthogonality appears with more than 3 hidden layers.  In Fig. \ref{Spiral_withT} we show that the use of a small temperature perturbation $\tau$ = 1e-6 speeds up training and slightly increases the test accuracy obtained for MLPs trained on the spiral data set. 
 
\begin{figure}[h]
    \centering
      \includegraphics[width=0.8\linewidth]{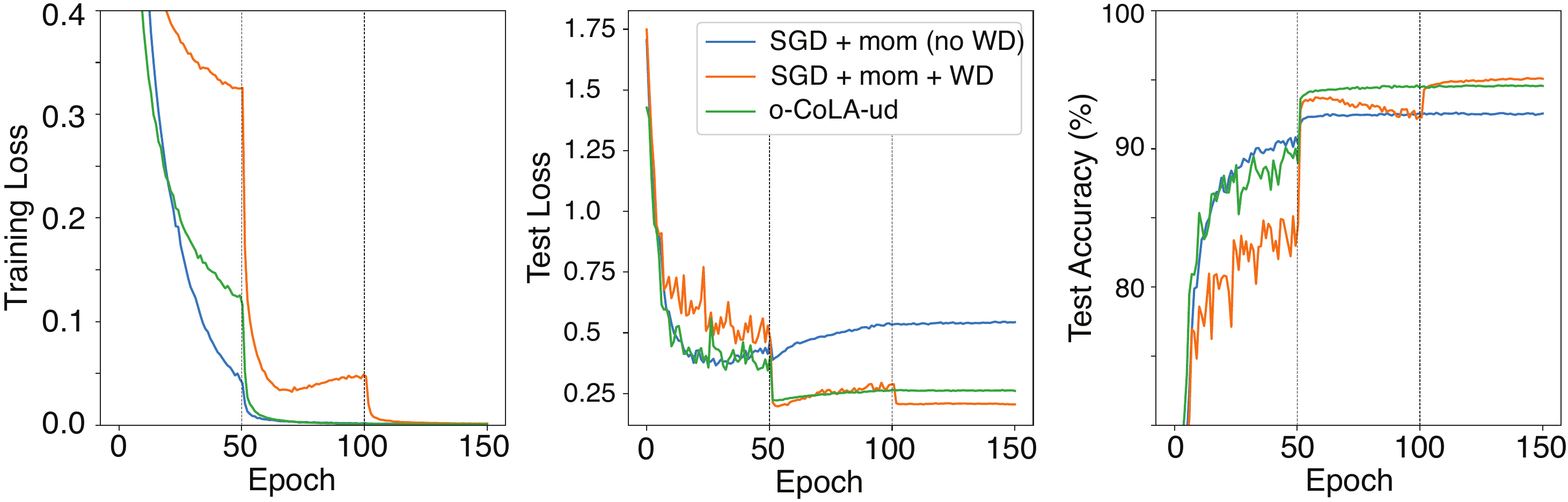} 
\vspace{-3mm}
    \caption{Train (left) \& test (middle) loss and test acc. (right) averaged over 5 runs of a ResNet-34 with BatchNorm trained using SGD-m vs. o-CoLA-ud with $\tau = 0$ on CIFAR-10. For SGD we initially use $h = 0.1$ and decay by a factor 10 every 50 epochs (indicated by the vertical black dotted lines). We set momentum = 0.9 and present results with and without WD. o-CoLA-ud (with $\gamma = 0.5$) did not use WD. Its learning rate was re-scaled to match the parameters of SGD-m and used the same LR schedule. The o-CoLA-ud method without weight decay strongly outperforms SGD-m without weight decay. \label{OG-OBA_Resnet}}
    
    \centering
       \includegraphics[width=1\linewidth]{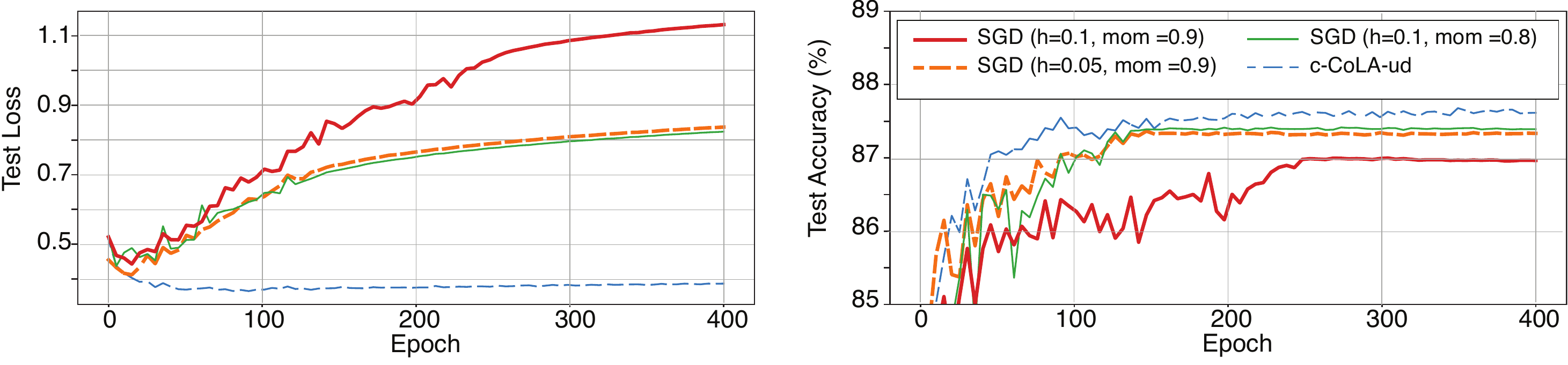}
\vspace{-8mm}
    \caption{Test loss (left) and test acc. (right) averaged over 5 runs of a 1000-node SHLP trained using SGD-m vs. c-CoLA-ud with $\tau = 0$ on Fashion-MNIST (batchsize 128, \# of training data samples reduced to 10K). Hyperparameters of c-CoLA-ud: $h = 0.3, \gamma = 1, r_0 = 0.05, r_1 = 0.1$. Due to the small training data set size both methods quickly reached 100\% training accuracy, but c-CoLA-ud is superior in its test loss and test accuracy.\label{CNet_FashionMNIST}}
\vspace{-0.45cm}
\end{figure}

For a ResNet-34 architecture with BatchNorm and learning rate (LR) decay on CIFAR-10 \cite{cifar10} data our underdamped orthogonal constrained method, o-CoLA-ud without weight decay (WD)  siginificantly outperforms SGD-m without WD (Fig. \ref{OG-OBA_Resnet}). In future work we will explore the nuances of combining orthogonality constraints with BatchNorm, residual connections and LR decay. 
Since o-CoLA outperforms SGD if no LR decay is used, we expect that with more tuning the use of WD can be completely removed by using orthogonality constraints (see also Fig. C\ref{CSGD_Resnet}).
\paragraph{Circle Constraints} We evaluate our circle constrained c-CoLA-ud method on the Fashion-MNIST data set \cite{FashionMNIST}. 
We reduce the amount of training data to 10K samples and use the remaining 60K samples 
as test data. 
c-CoLA-ud clearly outperforms SGD-m in terms of both test accuracy and test loss for a 1000-node single hidden layer perceptron (see Fig. \ref{CNet_FashionMNIST}). The lower test loss of c-CoLA-ud is maintained during training and the method shows no signs of overfitting, thus eliminating the need for early stopping. 
Even with weight decay, SGD-m is outperformed by its constrained counterpart (see Appx. \ref{tableFashionMNIST}). We also show that a small transformer \cite{Transformer} with 2 encoder layers (each with 2-head self-attention and a 200-node feed-forward network) trained using c-CoLA-ud achieves a lower validation loss on NLP datasets than its unconstrained counterpart, SGD-m (see Table \ref{NLPexp}).
\newpage

\begin{table}[h]
\vspace{-0.35cm}
\begin{tabular}{c|c|c|c|c|c|c|c} 
     &  & \multicolumn{3}{c|}{SGD $h$ = 0.1} & \multicolumn{3}{c}{SGD $h$ = 0.2} \\ 
      \textit{Data sets} & c-CoLA-ud & $mom=0.7$ & 0.8 & 0.9 & $mom=0.7$ & 0.8 & 0.9 \\ \hline 
      Penn Treebank & $\textbf{4.81}$ & 4.87 &
4.83 & 4.84 & 4.83 & 4.83 & 4.83 \\ \hline
      Wikitext-2 & $\textbf{5.09}$ & 5.13 & 5.13 & 5.13 & 5.13 & 5.14 & 5.13 \\
\end{tabular} \vspace*{-1mm}
\caption{Minimum validation loss on Penn Treebank data (batchsize 1024) \cite{PennTreebank} and Wikitext-2 (batchsize 128) \cite{Wikitext2} using a transformer trained using c-CoLA-ud (with $\tau = 0$) or SGD-m. Hyperparameters c-CoLA-ud: $h = 0.4, r = 0.5, r_L = 0.1, r_N = 1, r_A = 1, \gamma = 0.5$ (Treebank) and $\gamma = 1$ (Wikitext-2), where the subscripts $L, N, A$ represent the radii belonging to the linear, norm and self- attention layers respectively. The transformer trained using o-CoLA-ud obtains lower validation losses on both datasets.}\label{NLPexp}
\end{table}


\clearpage
\section*{Acknowledgements}
The authors wish to thank Gabriel Stoltz and Tony Lelièvre for helpful discussions on constrained
SDEs. Benedict Leimkuhler is a fellow of the Alan Turing Institute which is supported by EPSRC
grant EP/N510129/1. Timothée Pouchon is supported by the Swiss National Science Foundation,
project P2ELP2\_188037. Tiffany Vlaar is supported by The Maxwell Institute Graduate School
in Analysis and its Applications, a Centre for Doctoral Training funded by the UK Engineering
and Physical Sciences Research Council (grant EP/L016508/01), the Scottish Funding Council,
Heriot-Watt University and the University of Edinburgh.

\bibliography{references}

\begin{thebibliography}{52}
\providecommand{\natexlab}[1]{#1}
\providecommand{\url}[1]{\texttt{#1}}
\expandafter\ifx\csname urlstyle\endcsname\relax
  \providecommand{\doi}[1]{doi: #1}\else
  \providecommand{\doi}{doi: \begingroup \urlstyle{rm}\Url}\fi

\bibitem[Arjovsky et~al.(2016)Arjovsky, Shah, and Bengio]{UnitaryEvolutionRNNs}
M.~Arjovsky, A.~Shah, and Y.~Bengio.
\newblock Unitary evolution recurrent neural networks.
\newblock In \emph{International Conference on Machine Learning}, pages
  1120--1128, 2016.

\bibitem[Bakry and {\'E}mery(1985)]{BaE85}
D.~Bakry and M.~{\'E}mery.
\newblock Diffusions hypercontractives.
\newblock In J.~Az{\'e}ma and M.~Yor, editors, \emph{S{\'e}minaire de
  Probabilit{\'e}s XIX 1983/84}, pages 177--206, Berlin, Heidelberg, 1985.
  Springer Berlin Heidelberg.
\newblock ISBN 978-3-540-39397-9.
\newblock \doi{10.1007/BFb0075847}.

\bibitem[Bakry et~al.(2013)Bakry, Gentil, and Ledoux]{BGL13}
D.~Bakry, I.~Gentil, and M.~Ledoux.
\newblock \emph{Analysis and geometry of {M}arkov diffusion operators}, volume
  348.
\newblock Springer Science \& Business Media, 2013.

\bibitem[Bansal et~al.(2018)Bansal, Chen, and Wang]{OrthReg_CNNs}
N.~Bansal, X.~Chen, and Z.~Wang.
\newblock Can we gain more from orthogonality regularizations in training deep
  {CNN}s?
\newblock In \emph{Proceedings of the 32nd International Conference on Neural
  Information Processing Systems}, pages 4266--4276. Curran Associates Inc.,
  2018.

\bibitem[Beskos et~al.(2013)Beskos, Pillai, Roberts, Sanz-Serna, and
  Stuart]{Beskos2013}
A.~Beskos, N.~Pillai, G.~Roberts, J.-M. Sanz-Serna, and A.~Stuart.
\newblock Optimal tuning of the hybrid {M}onte {C}arlo algorithm.
\newblock \emph{Bernoulli}, 19\penalty0 (5A):\penalty0 1501--1534, 2013.

\bibitem[Bhattacharya(1982)]{Bha82}
R.~N. Bhattacharya.
\newblock On the functional central limit theorem and the law of the iterated
  logarithm for {M}arkov processes.
\newblock \emph{Zeitschrift f{\"u}r Wahrscheinlichkeitstheorie und verwandte
  Gebiete}, 60\penalty0 (2):\penalty0 185--201, 1982.
\newblock \doi{10.1007/BF00531822}.

\bibitem[Bou-Rabee and Sanz-Serna(2018)]{BRSS2017}
N.~Bou-Rabee and J.M. Sanz-Serna.
\newblock Geometric integrators and the {H}amiltonian {M}onte {C}arlo method.
\newblock \emph{Acta Numerica}, 27:\penalty0 113--206, 2018.
\newblock \doi{10.1017/S0962492917000101}.

\bibitem[Brock et~al.(2017)Brock, Lim, Ritchie, and Weston]{NeuralPhotoEditing}
A.~Brock, T.~Lim, J.~M. Ritchie, and N.~J. Weston.
\newblock Neural photo editing with introspective adversarial networks.
\newblock In \emph{5th International Conference on Learning Representations,
  {ICLR} 2017, Toulon, France, April 24-26, 2017, Conference Track
  Proceedings}. OpenReview.net, 2017.

\bibitem[Cheng et~al.(2017)Cheng, Chatterji, Bartlett, and Jordan]{Cheng2017}
X.~Cheng, N.~S. Chatterji, P.~L. Bartlett, and M.~I. Jordan.
\newblock {U}nderdamped {L}angevin {MCMC}: A non-asymptotic analysis.
\newblock \emph{arXiv:1707.03663}, 2017.

\bibitem[d'Ascoli et~al.(2019)d'Ascoli, Sagun, Bruna, and Biroli]{haystack}
S.~d'Ascoli, L.~Sagun, J.~Bruna, and G.~Biroli.
\newblock Finding the needle in the haystack with convolutions: on the benefits
  of architectural bias.
\newblock \emph{NeurIPS}, 2019.

\bibitem[Faou and Leli{\`e}vre(2009)]{FaL09}
E.~Faou and T.~Leli{\`e}vre.
\newblock Conservative stochastic differential equations: {M}athematical and
  numerical analysis.
\newblock \emph{Mathematics of computation}, 78\penalty0 (268):\penalty0
  2047--2074, 2009.
\newblock \doi{10.1090/S0025-5718-09-02220-0}.

\bibitem[Graham and Storkey(2017)]{GS2017}
M.~Graham and A.~Storkey.
\newblock Asymptotically exact inference in differentiable generative models.
\newblock In \emph{Proceedings of the 20th International Conference on
  Artificial Intelligence and Statistics}, volume~54, pages 499--508, 2017.

\bibitem[He et~al.(2015)He, Zhang, Ren, and Sun]{Pytorchinit}
K.~He, X.~Zhang, S.~Ren, and J.~Sun.
\newblock Delving deep into rectifiers: {S}urpassing human-level performance on
  {I}magenet classification.
\newblock In \emph{Proceedings of the IEEE international conference on computer
  vision}, pages 1026--1034, 2015.

\bibitem[Hoerl and Kennard(1970)]{ridge}
A.~Hoerl and R.~Kennard.
\newblock Ridge regression: {B}iased estimation for nonorthogonal problems.
\newblock \emph{Technometrics}, 12:\penalty0 55--67, 1970.
\newblock \doi{10.1080/00401706.1970.10488634}.

\bibitem[Huang et~al.(2018)Huang, Liu, Lang, Wei~Yu, and
  Li]{CNN_stiefelmanifold}
L.~Huang, X.~Liu, B.~Lang, A.~Wei~Yu, and B.~Li.
\newblock Orthogonal weight normalization: Solution to optimization over
  multiple dependent stiefel manifolds in deep neural networks.
\newblock In \emph{Thirty-Second AAAI Conference on Artificial Intelligence},
  2018.

\bibitem[Ioffe and Szegedy(2015)]{BatchNorm}
S.~Ioffe and C.~Szegedy.
\newblock Batch normalization: Accelerating deep network training by reducing
  internal covariate shift.
\newblock In \emph{International Conference on Machine Learning}, pages
  448--456, 2015.

\bibitem[Jia et~al.(2019)Jia, Li, Wen, Liu, and Tao]{OrthDNN}
K.~Jia, S.~Li, Y.~Wen, T.~Liu, and D.~Tao.
\newblock Orthogonal deep neural networks.
\newblock \emph{IEEE Transactions on Pattern Analysis and Machine
  Intelligence}, 2019.
\newblock \doi{10.1109/TPAMI.2019.2948352}.

\bibitem[Kipnis and Varadhan(1986)]{KiV86}
C.~Kipnis and S.~R.~S. Varadhan.
\newblock Central limit theorem for additive functionals of reversible {M}arkov
  processes and applications to simple exclusions.
\newblock \emph{Communications in Mathematical Physics}, 104\penalty0
  (1):\penalty0 1--19, 1986.
\newblock \doi{10.1007/BF01210789}.

\bibitem[Krizhevsky and Hinton(2009)]{cifar10}
A.~Krizhevsky and G.~Hinton.
\newblock Learning multiple layers of features from tiny images.
\newblock 2009.

\bibitem[Lee(2018)]{Lee18}
J.~M. Lee.
\newblock \emph{Introduction to {R}iemannian manifolds}, volume~2.
\newblock Springer, 2018.

\bibitem[Leimkuhler and Matthews(2016)]{LeM16}
B.~Leimkuhler and C.~Matthews.
\newblock Efficient molecular dynamics using geodesic integration and
  solvent--solute splitting.
\newblock \emph{Proceedings of the Royal Society A: Mathematical, Physical and
  Engineering Sciences}, 472\penalty0 (2189):\penalty0 20160138, 2016.
\newblock \doi{10.1098/rspa.2016.0138}.

\bibitem[Leimkuhler and Reich(2004)]{LeR04}
B.~Leimkuhler and S.~Reich.
\newblock \emph{Simulating {H}amiltonian dynamics}, volume~14.
\newblock Cambridge university press, 2004.

\bibitem[Leimkuhler et~al.(2016)Leimkuhler, Matthews, and Stoltz]{LMS16}
B.~Leimkuhler, C.~Matthews, and G.~Stoltz.
\newblock The computation of averages from equilibrium and nonequilibrium
  {L}angevin molecular dynamics.
\newblock \emph{IMA Journal of Numerical Analysis}, 36\penalty0 (1):\penalty0
  13--79, 2016.
\newblock \doi{10.1093/imanum/dru056}.

\bibitem[Leimkuhler et~al.(2019)Leimkuhler, Matthews, and Vlaar]{LMV}
B.~Leimkuhler, C.~Matthews, and T.~Vlaar.
\newblock {P}artitioned integrators for thermodynamic parameterization of
  neural networks.
\newblock \emph{Foundations of Data Science}, 1\penalty0 (4):\penalty0
  457--489, 2019.
\newblock \doi{10.3934/fods.2019019}.

\bibitem[Leli{\`e}vre and Stoltz(2016)]{LelS16}
T.~Leli{\`e}vre and G.~Stoltz.
\newblock Partial differential equations and stochastic methods in molecular
  dynamics.
\newblock \emph{Acta Numerica}, 25:\penalty0 681--880, 2016.
\newblock \doi{10.1017/S0962492916000039}.

\bibitem[Leli{\`e}vre et~al.(2010)Leli{\`e}vre, Stoltz, and Rousset]{LSR10}
T.~Leli{\`e}vre, G.~Stoltz, and M.~Rousset.
\newblock \emph{{F}ree energy computations: {A} mathematical perspective}.
\newblock Imperial College Press, 2010.
\newblock ISBN 9781848162488.

\bibitem[Leli{\`e}vre et~al.(2012)Leli{\`e}vre, Rousset, and Stoltz]{LRS12}
T.~Leli{\`e}vre, M.~Rousset, and G.~Stoltz.
\newblock Langevin dynamics with constraints and computation of free energy
  differences.
\newblock \emph{Mathematics of computation}, 81\penalty0 (280):\penalty0
  2071--2125, 2012.
\newblock \doi{10.1090/S0025-5718-2012-02594-4}.

\bibitem[Leli{\`e}vre et~al.(2020)Leli{\`e}vre, Stoltz, and Zhang]{LSZ2020}
T.~Leli{\`e}vre, G.~Stoltz, and W.~Zhang.
\newblock Multiple projection {MCMC} algorithms on submanifolds.
\newblock \emph{arXiv:2003.09402}, 2020.

\bibitem[Marcus et~al.(1993)Marcus, Santorini, and Marcinkiewicz]{PennTreebank}
M.~P. Marcus, B.~Santorini, and M.~A. Marcinkiewicz.
\newblock Building a large annotated corpus of {E}nglish: The {P}enn
  {T}reebank.
\newblock \emph{Computational Linguistics}, 19\penalty0 (2):\penalty0 313--330,
  1993.

\bibitem[Merity et~al.(2017)Merity, Xiong, Bradbury, and Socher]{Wikitext2}
S.~Merity, C.~Xiong, J.~Bradbury, and R.~Socher.
\newblock Pointer sentinel mixture models.
\newblock In \emph{5th International Conference on Learning Representations,
  {ICLR} 2017, Toulon, France, April 24-26, 2017, Conference Track
  Proceedings}. OpenReview.net, 2017.

\bibitem[Neyshabur et~al.(2015)Neyshabur, Tomioka, and Srebro]{neyshabur2015}
B.~Neyshabur, R.~Tomioka, and N.~Srebro.
\newblock In search of the real inductive bias: On the role of implicit
  regularization in deep learning.
\newblock In Yoshua Bengio and Yann LeCun, editors, \emph{3rd International
  Conference on Learning Representations, {ICLR} 2015, San Diego, CA, USA, May
  7-9, 2015, Workshop Track Proceedings}, 2015.

\bibitem[Pascanu et~al.(2013)Pascanu, Mikolov, and Bengio]{PascanuRNNs}
R.~Pascanu, T.~Mikolov, and Y.~Bengio.
\newblock On the difficulty of training recurrent neural networks.
\newblock In \emph{International conference on machine learning}, pages
  1310--1318, 2013.

\bibitem[Paszke et~al.(2017)Paszke, Gross, Chintala, Chanan, Yang, DeVito, Lin,
  Desmaison, Antiga, and Lerer]{Pytorch}
A.~Paszke, S.~Gross, S.~Chintala, G.~Chanan, E.~Yang, Z.~DeVito, Z.~Lin,
  A.~Desmaison, L.~Antiga, and A.~Lerer.
\newblock Automatic differentiation in {P}y{T}orch.
\newblock 2017.

\bibitem[Pavliotis(2014)]{Pav14}
G.~A. Pavliotis.
\newblock \emph{Stochastic processes and applications: diffusion processes, the
  {F}okker-{P}lanck and {L}angevin equations}, volume~60.
\newblock Springer, 2014.

\bibitem[Pennington et~al.(2017)Pennington, Schoenholz, and
  Ganguli]{Pennington2017}
J.~Pennington, S.~Schoenholz, and S.~Ganguli.
\newblock Resurrecting the sigmoid in deep learning through dynamical isometry:
  theory and practice.
\newblock In \emph{Advances in Neural Information Processing Systems}, pages
  4785--4795, 2017.

\bibitem[Pennington et~al.(2018)Pennington, Schoenholz, and
  Ganguli]{Pennington2018}
J.~Pennington, S.~Schoenholz, and S.~Ganguli.
\newblock The emergence of spectral universality in deep networks.
\newblock In \emph{International Conference on Artificial Intelligence and
  Statistics}, pages 1924--1932, 2018.

\bibitem[Rodr\'{i}guez et~al.(2017)Rodr\'{i}guez, Gonz\`{a}lez, Cucurull,
  Gonfaus, and Roca]{Rodriguez2017}
P.~Rodr\'{i}guez, J.~Gonz\`{a}lez, G.~Cucurull, J.~M. Gonfaus, and X.~Roca.
\newblock Regularizing cnns with locally constrained decorrelations.
\newblock In \emph{5th International Conference on Learning Representations,
  {ICLR} 2017, Toulon, France, April 24-26, 2017, Conference Track
  Proceedings}. OpenReview.net, 2017.

\bibitem[Santurkar et~al.(2018)Santurkar, Tsipras, Ilyas, and Madry]{BNhelpopt}
S.~Santurkar, D.~Tsipras, A.~Ilyas, and A.~Madry.
\newblock How does batch normalization help optimization?
\newblock In \emph{Advances in Neural Information Processing Systems}, pages
  2483--2493, 2018.

\bibitem[Saxe et~al.(2013)Saxe, McClelland, and Ganguli]{Saxe2013}
A.~M. Saxe, J.~L. McClelland, and S.~Ganguli.
\newblock Exact solutions to the nonlinear dynamics of learning in deep linear
  neural networks.
\newblock \emph{arXiv:1312.6120}, 2013.

\bibitem[Srebro and Shraibman(2005)]{maxnorm}
N.~Srebro and A.~Shraibman.
\newblock Rank, trace-norm and max-norm.
\newblock In \emph{International Conference on Computational Learning Theory},
  pages 545--560. Springer, 2005.
\newblock \doi{10.1007/11503415_37}.

\bibitem[Srivastava et~al.(2014)Srivastava, Hinton, Krizhevsky, Sutskever, and
  Salakhutdinov]{dropout2}
N.~Srivastava, G.E. Hinton, A.~Krizhevsky, I.~Sutskever, and R.~Salakhutdinov.
\newblock Dropout: a simple way to prevent neural networks from overfitting.
\newblock \emph{The journal of machine learning research}, 15\penalty0
  (1):\penalty0 1929--1958, 2014.

\bibitem[Tibshirani(1996)]{LASSO1}
R.~Tibshirani.
\newblock Regression shrinkage and selection via the lasso.
\newblock \emph{Journal of the Royal Statistical Society: Series B
  (Methodological)}, 58\penalty0 (1):\penalty0 267--288, 1996.
\newblock \doi{10.1111/j.2517-6161.1996.tb02080.x}.

\bibitem[Vaswani et~al.(2017)Vaswani, Shazeer, Parmar, Uszkoreit, Jones, Gomez,
  Kaiser, and Polosukhin]{Transformer}
A.~Vaswani, N.~Shazeer, N.~Parmar, J.~Uszkoreit, L.~Jones, A.~N. Gomez, Ł.
  Kaiser, and I.~Polosukhin.
\newblock Attention is all you need.
\newblock In \emph{Advances in Neural Information Processing Systems}, pages
  5998--6008, 2017.

\bibitem[Villani(2009)]{Vil09}
C.~Villani.
\newblock Hypocoercivity.
\newblock \emph{Memoirs of the American Mathematical Society}, 202\penalty0
  (950), 2009.

\bibitem[Vorontsov et~al.(2017)Vorontsov, Trabelsi, Kadoury, and
  Pal]{RNNsLongTermDependencies}
E.~Vorontsov, C.~Trabelsi, S.~Kadoury, and C.~Pal.
\newblock On orthogonality and learning recurrent networks with long term
  dependencies.
\newblock In \emph{Proceedings of the 34th International Conference on Machine
  Learning-{V}olume 70}, pages 3570--3578. JMLR. org, 2017.

\bibitem[Wenzel et~al.(2020)Wenzel, Roth, Veeling, Swiatkowski, Tran, Mandt,
  Snoek, Salimans, Jenatton, and Nowozin]{coldposterior}
F.~Wenzel, K.~Roth, B.~S. Veeling, J.~Swiatkowski, L.~Tran, S.~Mandt, J.~Snoek,
  T.~Salimans, R.~Jenatton, and S.~Nowozin.
\newblock How good is the {B}ayes posterior in deep neural networks really?
\newblock \emph{arXiv:2002.02405}, 2020.

\bibitem[Williams(1995)]{LASSO2}
P.~Williams.
\newblock Bayesian regularization and pruning using a laplace prior.
\newblock \emph{Neural computation}, 7\penalty0 (1):\penalty0 117--143, 1995.
\newblock \doi{10.1162/neco.1995.7.1.117}.

\bibitem[Xiao et~al.(2017)Xiao, Rasul, and Vollgraf]{FashionMNIST}
H.~Xiao, K.~Rasul, and R.~Vollgraf.
\newblock Fashion-{MNIST}: a novel image dataset for benchmarking machine
  learning algorithms.
\newblock \emph{arXiv:1708.07747}, 2017.

\bibitem[Xiao et~al.(2018)Xiao, Bahri, Sohl-Dickstein, Schoenholz, and
  Pennington]{10000CNN}
L.~Xiao, Y.~Bahri, J.~Sohl-Dickstein, S.~Schoenholz, and J.~Pennington.
\newblock Dynamical isometry and a mean field theory of {CNN}s: {H}ow to train
  10,000-layer vanilla convolutional neural networks.
\newblock In \emph{International Conference on Machine Learning}, pages
  5393--5402, 2018.

\bibitem[Xie et~al.(2017)Xie, Xiong, and Pu]{BeyondGoodInit}
D.~Xie, J.~Xiong, and S.~Pu.
\newblock All you need is beyond a good init: Exploring better solution for
  training extremely deep convolutional neural networks with orthonormality and
  modulation.
\newblock In \emph{Proceedings of the IEEE Conference on Computer Vision and
  Pattern Recognition}, pages 6176--6185, 2017.

\bibitem[Yao et~al.(2019)Yao, Gholami, Keutzer, and Mahoney]{PyHessian}
Z.~Yao, A.~Gholami, K.~Keutzer, and M.~Mahoney.
\newblock Py{H}essian: Neural networks through the lens of the {H}essian.
\newblock \emph{arXiv:1912.07145}, 2019.

\bibitem[Zappa et~al.(2018)Zappa, Holmes-Cerfon, and Goodman]{ZHG2018}
E.~Zappa, M.~Holmes-Cerfon, and J.~Goodman.
\newblock {M}onte {C}arlo on manifolds: {S}ampling densities and integrating
  functions.
\newblock \emph{Communications on Pure and Applied Mathematics}, 71\penalty0
  (12):\penalty0 2609--2647, 2018.
\newblock \doi{10.1002/cpa.21783}.

\end{thebibliography}
\newpage
\appendix
\textbf{Overview of the provided supplementary material:} \\ \ \\
Appendix \ref{sec:supplementSDEs}: Provides the results necessary to establish exponential convergence to equilibrium of constrained overdamped Langevin dynamics \eqref{eq:constrainedOverdampedLangevin}. \\ \ \\ 
Appendix \ref{sec:discretization}: Provides discretization schemes and implementation details for our constrained training algorithms. The discretization schemes for a general constraint are described in Appendix \ref{sec:generalCoLAod} for overdamped Langevin dynamics and in \ref{sec:generalCoLAud} for underdamped Langevin dynamics. Our c-CoLA circle constrained algorithm is discussed in Appendix \ref{sec:discr_OLcircle} (overdamped) and \ref{sec:discr_ULcircle} (underdamped). Appendix \ref{sec:OL_orthogonalConstraint} and \ref{sec:UL_orthogonalConstraint} are reserved for our o-CoLA, orthogonality constraint Langevin dynamics, algorithm (overdamped and underdamped, respectively).  \\ \ \\ 
Appendix \ref{sec:NumericsAppx}: Provides further numerical implementation details and results for our constrained methods.

\newpage
\section{Theory of constrained overdamped Langevin dynamics}	\label{sec:supplementSDEs}

We present here the details of the theory summarized in Sec. \ref{sec:constrainedSDEs}.	
In particular, we provide the key results and suitable references to establish the exponential convergence to equilibrium of
constrained overdamped Langevin dynamics \eqref{eq:constrainedOverdampedLangevin}.

In the first part (Sec. \ref{sec:proofUnderlyingSDE}), we derive the underlying SDE associated with \eqref{eq:constrainedOverdampedLangevin}, its generator and
the invariant measure $\nu_\manifold$ defined as
\begin{equation}	\label{eq:OL_invariantMeasure}
\d\nu_\manifold = Z^{-1} e^{-\beta V(q)} \, \d\sigma_\manifold,
\qquad
Z = \int_\manifold e^{-\beta V(q)} \, \d\sigma_\manifold,
\end{equation}
where $\sigma_\manifold$ is the surface measure on $\manifold$. Ergodicity ensures that averages of observables with respect to $\nu_\manifold$ can be approximated by time averages of trajectories of \eqref{eq:constrainedOverdampedLangevin}: for all test function $\phi\in\fcnC^\infty_c(\manifold)$ 
\begin{equation}	\label{eq:ergodicity}
\lim\limits_{T\to\infty}\intMean{\phi}_T = 
\intMean{\phi}_{\nu_{\manifold}}
\quad \text{for a.e. }q_0\in\manifold,
\qquad
\intMean{\phi}_T:= \frac{1}{T} \int_0^T \phi(q_t)  \,\d t
,\quad
\intMean{\phi}_{\nu_{\manifold}}:=\int_{\manifold} \phi(q) \,\d\nu_{\manifold}(q)
.
\end{equation}

Next, in Sec. \ref{sec:proofPoincare} we present the Poincar\'e inequality on a manifold, which holds under a curvature-dimension assumption: there exists $\rho>0$ such that
\begin{equation} \label{eq:assumptionV}
C\!D(\rho,\infty):\qquad
\quad\mathrm{Ric}_\frakg + \beta \nabla_\frakg^2 V 
 \geq \rho \frakg,
\end{equation}
in the sense of symmetric matrices.
The terms in \eqref{eq:assumptionV} rely on the structure of $\manifold$ as a Riemannian manifold:
$\frakg$ is the Riemannian metric, 
$\mathrm{Ric}_\frakg$ is the Ricci curvature tensor and
$\nabla_\frakg^2 V$ is the Hessian of $V$ on the manifold.
Under \eqref{eq:assumptionV} the following result holds \cite{BGL13}.
\vspace*{-0.2cm}
\begin{theorem}	\label{thm:OL_poincare}
Assume that there exists $\rho>0$ and $N>n$ such that $CD(\rho,N)$ holds.
Then $\nu_\manifold$ satisfies a Poincaré inequality:
there exists a constant $L>0$ such that
\begin{equation}	\label{eq:OL_poincare}
\int_{\manifold} \big|\phi(q) -\intMean{\phi}_{\nu_\manifold} \big|^2 \,\d\nu_\manifold(q) 
	\leq \frac{1}{2L} \int_{\manifold} \big|\Pi(q)\nabla_q \phi(q)\big|^2 \,\d\nu_\manifold(q)
\qquad \forall \phi\in H^1(\nu_\manifold),
\end{equation}
where $\Pi(q)$ is the projection onto the cotangent space $T_q^*\manifold$ \eqref{eq:OL_definitionPi} and
$H^1(\nu_\manifold)$ is the space of functions with square $\nu_\manifold$-integrable gradients \eqref{eq:defH1}.
\end{theorem}
Consequences of Theorem \ref{thm:OL_poincare} are exponential convergence and a central limit theorem (CLT) for the convergence in \eqref{eq:ergodicity}.
\vspace*{-0.2cm}
\begin{corollary}	\label{cor:geometricErgodicityAndClt}
If \eqref{eq:assumptionV} holds then
\begin{equation}	\label{eq:geometricErgodicity}
\int_\manifold \big| \E( \phi(q_t) \mid q_0 ) - \intMean{\phi}_{\nu_\manifold} \big|^2 \,\d\nu_{\manifold}(q_0)
	 \leq C(\phi)e^{-2L/\beta t}
\qquad \forall \phi\in H^1(\nu_\manifold),
\end{equation}
where $C(\phi)$ depends only on $\phi$.
Furthermore we have the following convergence in law:
\[
\sqrt{T}\big( \intMean{\phi}_T - \intMean{\phi}_{\nu_\manifold} \big) \to \mathcal{N} ( 0 , \sigma_\phi^2 )
\quad \text{as }T\to\infty,
\]
where the asymptotic variance $\sigma_\phi^2$ is bounded as
$\sigma_\phi^2 \leq  \frac{\beta}{L} \int_{\manifold} \big|\phi -\intMean{\phi}_{\nu_\manifold} \big|^2 \,\d\nu_\manifold$.
\end{corollary}
Appx. \ref{sec:proofCorollaries} is dedicated to using the Poincar\'e inequality to proving this. 

In $\R^n$, assumption \eqref{eq:assumptionV} is equivalent to  convexity of $V$, which is known to be too strong a requirement (a confining assumption is sufficient, see e.g. \cite{LelS16}).
Although \eqref{eq:assumptionV} can certainly be weakened, the above results ensure that, provided the curvature of the manifold is well behaved, sampling on $\manifold$ has similar properties as on a flat space.

\subsubsection*{Notation}
We collect here additional notation needed for this discussion.

Given a measure $\mu$ in a space $E\subset \R^d$, we associate the space of square integrable functions
\begin{equation*}	
\Ld(\mu)= \big\{ \phi:E\to\R \text{ measurable } : \int_{E} |\phi|^2\,\d \mu < \infty \big\}.
\end{equation*}
Equipped with the inner product and associated norm
\[
\ps{ \phi, \psi }_{\mu} = \int_{E} \phi \psi\,\d \mu,
\qquad
\norm{\phi}_{\Ld(\mu)} = \sqrt{\psbig{ \phi, \phi }},
\]
$\Ld(\mu)$ is a Hilbert space.
We further define the subspace  $\Ld(\mu)$ of functions with zero mean by
\begin{equation}
\Ld_0(\mu) = \big\{ \phi\in \Ld(\mu) : \intMean{\phi}_\mu = 0\big\},
\quad
\intMean{\phi}_\mu = \int_{E} \phi \d \mu,
\end{equation}
as well as the space of functions with square integrable gradient
\begin{equation}	\label{eq:defH1}
H^1(\mu) = \big\{ \phi\in \Ld(\mu): \partial_i\phi\in \Ld(\mu) \quad 1\leq i\leq d\big\}.
\end{equation}

For the constraint $g:\R^d\to\R^m$,
we denote the Jacobian matrix as $G(q) = \nabla_q^Tg(q)$ and denote its right pseudo-inverse by $G^{+} = G^T(GG^T)^{-1}$ ($GG^T$ is invertible if $G$ has full row rank).
We verify that the map
\begin{equation}	\label{eq:OL_definitionPi}
\Pi:\R^d\to\R^{d \times d}, 
\quad q\mapsto \Pi(q) = I_d - G^+(q)G(q),
\end{equation}
defines for each $q$ the orthogonal projection onto the cotangent space $T_q^*\manifold$.
\[
\Pi_q=\Pi(q) : \R^d\to\R^d,
\quad p\mapsto \Pi(q)p .
\]
In particular, for all $q$ we have $\Pi_q p\in T_q^*\manifold$ and the matrix $\Pi_q$ is symmetric and idempotent:
(i.e., $\Pi_q^T=\Pi_q$ and $\Pi_q^2=\Pi_q$).

\subsection{The underlying SDE and the invariant measure}	\label{sec:proofUnderlyingSDE}

Although presented differently, the results of this section follow closely the treatment of this issue presented in \cite[Chap. 3]{LSR10}.

We define the mean curvature of the manifold as the vector valued function
\begin{equation}	\label{eq:OL_definitionH}
\mathcal{H}:\R^d\to\R^d,
\quad q\mapsto  \big(\mathcal{H}(q)\big)_i = \Pi_{jk}(q)\partial_j \Pi_{ik}(q) \quad 1\leq i\leq d,
\end{equation}
where $\Pi(q):\R^d\to\R^d$ is the projection onto the cotangent space defined in \eqref{eq:OL_definitionPi}.
We then establish the following result (proved below).
\begin{lemma}	\label{lem:OL_augmentedDynamics}
	The constrained system \eqref{eq:constrainedOverdampedLangevin} can be rewritten as the following SDE in $\R^d$
	\begin{equation}	\label{eq:OL_augmentedDynamics}
	\d q_t = -\Pi(q_t) \nabla V(q_t)\d t +  \sqrt{2\beta^{-1}}\,\Pi(q_t)\d \mathcal{W}_t  + \beta^{-1}\mathcal{H}(q_t) \,\d t.
	\end{equation}
\end{lemma}


The uniqueness of the invariant measure of \eqref{eq:OL_augmentedDynamics} and the resulting ergodicity result \eqref{eq:ergodicity} are proved in \cite[Prop. 3.20]{LSR10} (the proof relies on the divergence theorem on manifolds).

The generator associated with \eqref{eq:OL_augmentedDynamics} is given by
\[
\calL = -\Pi(q) \nabla V(q)\cdot \nabla + \beta^{-1}\mathcal{H}(q) \cdot \nabla + \beta^{-1}\Pi(q):\nabla^2.
\]
We verify that $\calL$ can be written in the following symmetric form 
\begin{equation}	\label{eq:OL_Lsymmetric}
\calL\psi 
=	\beta^{-1}\Div_\manifold(\nabla_\manifold\psi) - \nabla_\manifold V(q) \cdot \nabla_\manifold\psi
=	\beta^{-1} e^{\beta V(q)} \Div_\manifold\big( e^{-\beta V(q)} \nabla_\manifold \psi\big),
\end{equation}
where we denote
$\nabla_\manifold \phi = \Pi\nabla\phi$
and 
$\Div_\manifold \psi = \nabla_\manifold\cdot\psi = \sum_{i,j=1}^d \Pi_{ij}\partial_j \psi_i$.
This expression directly implies that $\calL$ is reversible with respect to $\nu_{\manifold}$:
\begin{equation}	\label{eq:OL_reversibility}
\psbig{ \calL\phi, \psi }_{\nu_\manifold} = -\beta^{-1} \psbig{ \nabla_\manifold \phi, \nabla_\manifold \psi }_{\nu_\manifold}
= \psbig{ \phi, \calL\psi }_{\nu_\manifold}.
\end{equation}


Thanks to this expression, we can prove that the measure $\nu_{\manifold}$ is indeed invariant for \eqref{eq:constrainedOverdampedLangevin}.
Let us introduce the forward Kolmogorov equation: given a test function $\phi\in\fcnC^\infty_c(\manifold)$
\[
\partial_t u(t,q) = \calL u(t,q) ~t\geq0,~q\in\manifold\qquad u(0,q) = \phi(q).
\]
The solution to this equation is verified to be
$u(t,q) = \E(\phi(q_t)\mid q_0=q)$ (see the Feynmann--Kac formula)
and is usually denoted as $u(t,q) = e^{t\calL}\phi(q)$.
The measure $\nu_\manifold$ is invariant if for any $t\geq 0$
$\int_\manifold u(t,q) \,\d \nu_\manifold(q) = \int_\manifold u(0,q) \,\d \nu_\manifold(q) = \intMean{\phi}_{\nu_\manifold}$.
This is easily verified thanks to \eqref{eq:OL_reversibility}:
\[
\frac{\d}{\d t} \int_\manifold u(t,q) \,\d \nu_\manifold(q)
=
\frac{\d}{\d t} \int_{\manifold} e^{t\calL}\phi(q) \,\d \nu_{\manifold}(q)
= \int_{\manifold} \calL e^{t\calL}\phi(q) \,\d \nu_{\manifold}(q)
= \psbig{ \calL e^{t\calL}\phi, \mathbf{1} }_{\nu_{\manifold}} = 0.
\]

\begin{proof} 
Let us write $\lambda_t$ as the It\^o process
\begin{equation}	\label{eq:lambdatAnsatz}
\d\lambda_t = \mu(q_t) \, \d t + \sigma(q_t) \, \d \mathcal{W}_t,
\end{equation}
where 
$\mu:\R^d\to\R^m$, $\sigma:\R^d\to\R^{m\times d}$
and $\mathcal{W}_t$ is the same Wiener process as in \eqref{eq:constrainedOverdampedLangevin}.
Using this expression in \eqref{eq:constrainedOverdampedLangevin} brings
\[
\d q_t = \big(-\nabla V(q_t)-G(q_t)^T\mu(q_t)\big)\,\d t + \big(\sqrt{2\beta^{-1}}I- G(q_t)^T\sigma(q_t)\big)\,\d \mathcal{W}_t,
\]
where we recall the notation for the Jacobian $G = \nabla_q^T g$.
Using It\^o formula we find 
\begin{equation}	\label{eq:dgqt}
0=\d g(q_t) = G(q_t)\,\d q + b_t\,\d t
= G(q_t)\big( -\nabla V(q_t) \,\d t+  \sqrt{2\beta^{-1}}\,\d \mathcal{W}_t  -G(q_t) ^T \d \lambda_t\big) + b_t\,\d t,
\end{equation}
where $b_t$ is the $d$-dimensional process defined as (omitting the dependence on $q_t$)
\begin{equation}	\label{eq:defbt}
\begin{aligned}
(b_t)_i &= \frac12 \big(\sqrt{2\beta^{-1}}I- G^T\sigma\big)\big(\sqrt{2\beta^{-1}}I- G^T\sigma\big)^T : \nabla^2 g_i
\\&= \beta^{-1}\Delta g_i
- \frac{\sqrt{2\beta^{-1}}}{2} \big(G^T\sigma+ \sigma^TG\big) :\nabla^2 g_i
+ \frac12 G^T\sigma\sigma^T G :\nabla^2 g_i.
\end{aligned}
\end{equation}
From \eqref{eq:dgqt} yields
\begin{equation}	\label{eq:expressionLambda}
\d \lambda_t = \big(G(q_t)G(q_t)^T\big)^{-1} G(q_t) \Big( -\nabla V(q_t) \,\d t + \sqrt{2\beta^{-1}} \,\d \mathcal{W}_t \Big) + \big(G(q_t)G(q_t)^T\big)^{-1} b_t \,\d t.
\end{equation}
Identifying with \eqref{eq:lambdatAnsatz} we find $\sigma(q) = \sqrt{2\beta^{-1}} (G^+(q))^T$, which used in \eqref{eq:defbt} yields
\[
(b_t)_i = \beta^{-1} \big(\Delta g_i - \big(G^T(G^+)^T+ G^+G\big) :\nabla^2 g_i + G^T(G^+)^T G^+ G :\nabla^2 g_i\big ).
\]
As $G^+G$ is symmetric and $GG^+= I_m$, we obtain
\begin{equation}	\label{eq:expressionbt}
(b_t)_i = \beta^{-1}\big(\Delta g_i - G^+G :\nabla^2 g_i\big) = \beta^{-1} \Pi : \nabla^2 g_i.
\end{equation}
Inserting \eqref{eq:expressionLambda} in \eqref{eq:constrainedOverdampedLangevin} brings
\begin{equation}	\label{eq:expressiondqt}
\d q_t = -\Pi(q_t) \nabla V(q_t)\d t +  \sqrt{2\beta^{-1}}\,\Pi(q_t)\d \mathcal{W}_t  - G^+(q_t) b_t \,\d t.
\end{equation}
To conclude the proof we require the following technical relations on the mean curvature vector
(\eqref{eq:expressionHwithdivPi} follows from a direct computation; the proof of \eqref{eq:equalityPiH} is direct but involved and can be found in \cite[Lemma 3.15]{LSR10}).
\begin{lemma}	\label{lem:relationsCurvature}
	The projection $\Pi$ and the vector $H$ defined in \eqref{eq:OL_definitionPi} and \eqref{eq:OL_definitionH} satisfy the following equalities
	\begin{subequations}
		\begin{flalign}	
		\mathcal{H} &= (I-\Pi)\nabla\cdot \Pi,	
		\label{eq:expressionHwithdivPi}
		\\
		\Pi:\nabla^2g_i &= -(G\mathcal{H})_i\qquad	1\leq i\leq d,	
		\label{eq:equalityPiH}
		\end{flalign}
	\end{subequations}
\end{lemma}
Equality \eqref{eq:expressionHwithdivPi} ensures that $\Pi \mathcal{H} = 0$.
Combining \eqref{eq:expressionbt} and \eqref{eq:equalityPiH} we can write $b_t = -\beta^{-1}G\mathcal{H}$.
Thanks to these relations and the definition of $\Pi$, we obtain
\[
-G^+b_t = \beta^{-1}G^+G\mathcal{H} =\beta^{-1}(I-\Pi)\mathcal{H} = \beta^{-1} \mathcal{H}.
\]
This equality combined with \eqref{eq:expressiondqt} proves \eqref{lem:OL_augmentedDynamics} and concludes the proof of Lemma \ref{lem:OL_augmentedDynamics}.
\end{proof}

\subsection{Poincar\'e inequality on a manifold}	\label{sec:proofPoincare}

Poincar\'e inequalities, also called spectral gap inequalities, form an important family of functional inequalities in the theory of Markov diffusion processes.
They are the simplest inequalities that provide results on the convergence to equilibrium.
Stronger results can be obtained with the family of log-Sobolev inequalities, which are at the center of the Bakry--\'Emery theory \cite{BaE85}.
We follow here closely the book \cite{BGL13} on this subject (more specifically §1.16.2 and sections 4.2, 4.8, C.6).
For the necessary terminology of  Riemannian manifolds we recommend the introductory textbook \cite{Lee18} (the literature on this topic is vast and contains many works of high quality).

As presented in \cite[Chap. 4]{BGL13}, a Poincar\'e inequality can be obtained as a consequence of a curvature-dimension condition.
For the sake of  presentation, we introduce this result in the setting of a weighted Riemannian manifold.
Let $(\calM,\frakg)$ be an $n$-dimensional Riemannian manifold, where $\frakg$ is the Riemannian metric.
We consider the diffusion operator
\[
\calL = \Delta_\frakg - \langle \nabla_\frakg W , \nabla_\frakg \cdot \rangle_\frakg ,
\]
where $\Delta_\frakg$ denotes the Laplace--Beltrami operator on the manifold $\calM$, $\nabla_\frakg$ denotes the Levi--Civita connection (covariant derivative) and $\langle\cdot,\cdot\rangle_\frakg$ denotes the Riemannian metric
($\langle X,Y\rangle_\frakg = \frakg(X,Y)$ for all vector fields $X,Y$).
We verify that the associated invariant measure is
$\d\mu = Z^{-1}e^{-W}\d\mu_\frakg$, where $\d\mu_\frakg$ is the Riemannian measure \cite[§1.11.3]{BGL13}.
For $N\in[n,\infty]$, we define the 2-tensor 
\[
\mathrm{Ric}_{N}(\calL) = \mathrm{Ric}_\frakg + \nabla_\frakg^2 W - \frac{1}{N-n} \d W\otimes \d W.
\]
where $\mathrm{Ric}_\frakg$ is the Ricci curvature $2$-tensor and
$\nabla_\frakg^2$ denotes the Hessian operator on $\calM$ (the case $N=n$ is considered only if $W$ is constant).
In this context, a curvature-dimension condition $C\!D(\rho,N)$ for $\rho\in\R$ and $N\geq n$ holds if and only if (see \cite[C.6]{BGL13})
\begin{equation}	\label{eq:CD}
C\!D(\rho,N):\qquad\mathrm{Ric}_{N}(\calL) \geq \rho \frakg,
\end{equation}
in the sense of symmetric $(0,2)$-tensors (covariant 2-tensors).
In the flat space $\calM = \R^n$, the condition $C\!D(\rho,\infty)$ reads $\nabla^2 W\geq \rho I$, which is nothing but the convexity of the potential $W$.
Under $C\!D(\rho,N)$, the measure $\mu$ is proved to satisfy a Poincar\'e inequality 
(in \cite{BGL13}, combine Thm 4.8.4 with the discussion in section C.6).
\begin{theorem} \label{thm:poincareManifold}
	\cite[Thm 4.8.4]{BGL13}
Under the curvature-dimension condition $C\!D(\rho,N)$ with $\rho>0$ and $N\geq n$, $N>1$, 
the measure $\mu$ satisfies the Poincar\'e inequality
\begin{equation}	\label{eq:OL_poincare_app2}
\mathrm{Var}_\mu(\phi) = \norm{\phi-\intMean{\phi}_\mu}_{\Ld(\mu)}^2
	\leq C_P\norm{ \nabla_\frakg\phi }_{\Ld(\mu)}^2
\qquad \forall \phi\in \Ld(\mu) \cap H^1(\mu),
\end{equation}
with constant $C_P = \frac{N-1}{\rho N}$.
\end{theorem}

As the tensor $\d W\otimes \d W$ is positive semi-definite,
we verify the monotonicity $\mathrm{Ric}_{N+M}(\calL)\geq \mathrm{Ric}_{N}(\calL)$ for any $M\geq 0$.
This implies in particular that $CD(\rho,N)\Rightarrow CD(\rho,\infty)$ for any $N\in[n,\infty]$.
Hence, among all choices of $N\geq n$, $CD(\rho,\infty)$ is the weaker condition.

Let us now consider this result in the context of the constraint manifold $\manifold$ in \eqref{eq:constraintManifold}.
We consider the space $\R^d$ with its Riemannian manifold structure given by the Euclidean metric 
$\bar\frakg (v,w)= v\cdot w$ for all $v,w\in\R^d$ (for all $q\in\R^d$, $p\in T_q\R^d$ is identified with $\R^d$ through a canonical isomorphism).
Assuming that $g$ is smooth and that $\nabla_q^Tg$ has everywhere full row-rank, $\manifold$ is a smooth embedded submanifold of $\R^d$ of dimension $n = d-m$ (see e.g. \cite[Cor. A.26]{Lee18}).
Furthermore, $\manifold$ is equipped with the metric induced by $\bar\frakg$:
for a local parameterization of $\psi:U\subset\manifold\to\R^d$, $\bar \frakg$ is given locally on $U$ by 
\begin{equation}	\label{eq:inducedMetric}
\bar\frakg 
	= \sum_{i=1}^d\sum_{j,k=1}^n 
		\frac{\partial \psi^i}{\partial {x^j}} \frac{\partial \psi^i}{\partial {x^k}}  \,\d x^j \d x^k
	= \big(\nabla_x \psi \nabla_x^T \psi\big)_{jk} \,\d x^j \d x^k .
\end{equation}
We now define the potential $W = \beta V|_\manifold$, where $V|_\manifold$ denotes the restriction of $V$ to $\manifold$.
Assumption \ref{eq:assumptionV} corresponds then to condition $C\!D(\rho,\infty)$ above.
Applying Theorem \ref{thm:poincareManifold} we obtain Poincar\'e's inequality on the constraint manifold $\manifold$.
We note that for a function $\phi$ defined on $\R^d$, the covariant derivative in $\R^d$ of $\phi|_{\manifold}$ on the manifold is the orthogonal projection of the directional derivative of $\phi$ (in the ambient manifold $\R^d$) onto the cotangent space: 
$\nabla_\frakg(\phi|_{\manifold})(q) = \Pi(q)\nabla_q \phi(q)$.
Furthermore, we note that the surface measure $\sigma_\manifold$ equals the Riemannian measure on the manifold
(compare \cite[Rem. 3.4]{LSR10} with \cite[Prop. 2.41]{Lee18} and \eqref{eq:inducedMetric}).
We thus obtain the result of Theorem \ref{thm:OL_poincare}
with constant $C_P = \frac1\rho = \frac{1}{2L}$.

\subsection{Exponential convergence to equilibrium and central limit theorem}	\label{sec:proofCorollaries}

Let us define the norm of a linear operator $\calA:\Ldznu\to\Ldznu$ as
\[
\norm{\calA}_{\mathcal{B}(\Ldznu)} = \sup_{\phi\in\Ldznu} \frac{\norm{\calA\phi}_{\Ldznu}}{\norm{\phi}_{\Ldznu}}.
\]
Denote $\bar{\phi} = \phi-\intMean{\phi}_{\nu_\manifold}\in\Ldznu$. The Poincar\'e inequality \eqref{eq:OL_poincare}, rewritten on the subspace $\Ldznu$, is as follows:
\begin{equation}	\label{eq:OL_poincare_app}
\norm{\bar\phi}_{\Ldznu}^2	\leq \frac{1}{2L}\norm{ \nabla_\manifold\bar\phi }_{\Ldznu}^2
\qquad \forall\bar\phi\in \Ldznu \cap H^1(\nu_\manifold).
\end{equation}
Using the reversibility of the measure \eqref{eq:OL_reversibility}, we can prove the following result (the proof follows the same lines as \cite[Prop. 2.3]{LelS16}, see also \cite[Thm 4.2.5]{BGL13}).
\begin{lemma} \label{lem:exponentialConvergence}
The measure $\nu_\manifold$ satisfies the Poincar\'e inequality \eqref{eq:OL_poincare_app} if and only if
\begin{equation}	\label{eq:exponentialConvergence}
\norm{e^{t\calL}}_{\mathcal{B}(\Ldznu)}
	\leq e^{-2\frac{L}{\beta}t}.
\end{equation}
\end{lemma}
Exponential convergence to equilibrium is then directly obtained from Lemma \ref{lem:exponentialConvergence}: 
\begin{equation}	\label{eq:exponentialConv_app}
\norm{e^{t\calL}\bar\phi}_{\Ldznu}
	\leq \norm{e^{t\calL}}_{\mathcal{B}(\Ldznu)} \norm{\bar\phi}_{\Ldznu} 
	\leq e^{-2\frac{L}{\beta}t} \norm{\bar\phi}_{\Ldznu}.
\end{equation}
This inequality implies \eqref{eq:geometricErgodicity} (note that $e^{t\calL} \intMean{\phi}_{\nu_\manifold} = \intMean{\phi}_{\nu_\manifold}$) and thus proves the first assertion of Corollary \ref{cor:geometricErgodicityAndClt}.

A consequence of the exponential convergence to equilibrium \eqref{eq:exponentialConv_app} is the following central limit theorem for time averages $\intMean{\phi}_T = \frac{1}{T}\int_{0}^T \phi(q_t) \,\d t$ (see also \cite{KiV86}).
\begin{theorem}	\label{thm:clt_app}
	\cite{Bha82}
	If \eqref{eq:exponentialConv_app} holds, then the following convergence in law is satisfied
	\[
	\sqrt{T}\big( \intMean{\phi}_T - \intMean{\phi}_{\nu_\manifold} \big) \to \mathcal{N} ( 0 , \sigma_\phi^2 )
	\quad \text{as }T\to\infty,
	\]
	where the asymptotic variance $\sigma_\phi^2$ is given by the formula
	$\sigma_\phi^2	= 2 \ps{\bar\phi, -\calL^{-1} \bar\phi}$
	with $\bar\phi = \phi-\intMean{\phi}_{\nu_\manifold}$.
\end{theorem}

To quantify the asymptotic variance, we use the following classical result.
\begin{lemma}	\label{lem:exponentialConvergence_App2}
	(e.g., \cite[Prop. 2.1]{LelS16})
If \eqref{eq:exponentialConvergence} holds, then the generator $\calL$ is invertible and the resolvent can be expressed as $-\calL^{-1} = \int_{0}^\infty e^{t\calL} \,\d t$ and satisfies the bound
$\norm{\calL^{-1}}_{\mathcal{B}(\Ldznu)} \leq \frac{\beta}{2L}$.
\end{lemma}

Using Lemma \ref{lem:exponentialConvergence_App2} and Cauchy--Schwartz inequality, the asymptotic variance in Theorem \ref{thm:clt_app} can thus be bounded as
\[
\sigma_\phi^2
	= 2 \int_{\manifold} \bar\phi (-\calL^{-1} \bar\phi) \,\d \nu_\manifold
	\leq 2 \norm{\calL^{-1}}_{\mathcal{B}(\Ldznu)} \norm{\bar\phi}_{\Ldznu}^2 
	\leq \frac{\beta}{L} \norm{\bar\phi}_{\Ldznu}^2.
\]
This estimate completes the proof of the second assertion of Corollary \ref{cor:geometricErgodicityAndClt}.

\section{Discretization of constrained Langevin dynamics}	\label{sec:discretization}

We present here the details of the  constrained training methods considered in this paper.
Both the overdamped \eqref{eq:constrainedOverdampedLangevin} and underdamped \eqref{eq:constrainedUnderdampedLangevin_sde} Langevin dynamics are discretized for the constraints presented in Section \ref{sec:constraints}.
We emphasize that the initialization of each given method must be done with care: the constrained parameters, the potential slack variable, as well as their momenta in the underdamped case, have to satisfy the constraint initially.

Recall the notation introduced in Section \ref{sec:constraints}:
$\theta\in\R^{|n|}$ is the vector of all the parameters of the model,
we consider the variable $q = (\theta,\xi)\in\R^{d}$, $d=|n|+n^\xi$,
where $\xi\in\R^{n^{\xi}}$ is a slack variable to enforce the potential inequality constraints.
The loss is extended $q=(\theta,\xi)$ as
$V(q) = \loss(\theta)$ (in particular $\nabla_\xi V = 0$) 
and constraints are given by a map $g:\R^d\to\R^m$.
The parameters are partitioned as $\theta = (\theta^u,\theta^c)$,
where $\theta^u\in\R^{n^u}$ are not involved in any constraint
while $\theta^c\in\R^{n^c}$ are.

\subsection{Discretization of constrained overdamped Langevin (general constraint)}\label{sec:generalCoLAod}

Following \cite[Chap. 3]{LSR10} a simple discretization of the constrained overdamped Langevin dynamics \eqref{eq:constrainedOverdampedLangevin} is given by the iteration
$q_n\in\manifold\mapsto q_{n+1}$ defined as
\begin{equation}
\label{eq:OLdiscretization1}
\begin{aligned}
&\bar{q}_{n+1} = q_n -\nabla_q V(q_n)h +  \sqrt{2\beta^{-1}h}\,R_n,  
\qquad q_{n+1}= \bar{q}_{n+1}-\nabla_q g(q_n) \lambda_n,	\\
&\text{where $\lambda_n\in\R^m$ is such that }g(q_{n+1})=0,
\end{aligned}
\end{equation}
where $R_n\sim N(0,I)$ is a vector of iid standard normal random variable.
The first step of \eqref{eq:OLdiscretization1}, $\bar{q}_{n+1}$,
is an Euler--Maruyama step for standard overdamped Langevin.
As $\bar{q}_{n+1}$ in $\R^d$ is generally not on the constrained manifold $\manifold$, the last term is present to project $\bar{q}_{n+1}$ back onto  $\manifold$, ensuring  $g(q_{n+1})=0$.
In particular, for the unconstrained parameter we have 
$\nabla_{\theta^u}^Tg ={0}_{m\times n^u}$ which implies that $\theta^u_{n+1} = \bar\theta^u_{n+1}$ is a standard EM step.

In general, projecting back onto the manifold $\manifold$, i.e., finding $\lambda_n$,  can be done using root-finding algorithms.
Nevertheless, for certain constraints $g$ the roots can be found explicitly.
This is the case for the circle constraint \eqref{eq:circleConstraint} 
(see Section \ref{sec:discr_OLcircle}). 
A potential weakness of method \eqref{eq:OLdiscretization1} is that the projection process can be guaranteed only for small enough step size $h$ (i.e. $\bar q_n$ must be close to $\manifold$).
Indeed, even for the circle constraint if $h$ is too large it might not be possible to project $\bar{q}_{n+1}$ back onto the circle following the direction $\nabla_q g(q_n)$.  See \cite{LSZ2020} for some discussion of methods to allow computation to be performed in the large timestep regime.

An alternative method is given by the iteration
$q_n\in\manifold\mapsto q_{n+1}\in\manifold$ defined as in \cite[Chap. 3]{LSR10}
\begin{equation}
\label{eq:OLdiscretization2}
\begin{aligned}
&\bar{q}_{n+1} = q_n -\nabla_q V(q_n)\d t +  \sqrt{2\beta^{-1}h}\,R_n,  
\qquad q_{n+1} = \bar{q}_{n+1}-\nabla_q g({q}_{n+1}) \lambda_n,	\\
&\text{where $\lambda_n\in\R^m$ is such that }g(q_{n+1})=0,
\end{aligned}
\end{equation}
where $R_n\sim N(0,I)$ is a vector of iid standard normal random variable.
The projection used in method \eqref{eq:OLdiscretization2} is in general more robust.
The circle constraint is a good illustration of this: while in \eqref{eq:OLdiscretization1} we project following an oblique direction, in \eqref{eq:OLdiscretization2} the projection is orthogonal and always exists (see Section \ref{sec:discr_OLcircle}).

\subsection{Discretization of constrained underdamped Langevin (general constraint)}\label{sec:generalCoLAud}

We next consider  the discretization of the constrained underdamped Langevin dynamics \eqref{eq:constrainedUnderdampedLangevin_sde} where we denote by $p = (p^u,p^c,p^\xi)\in\R^{n^u+n^c+n^\xi}$ the momenta associated with the configuration $q = (\theta^u,\theta^c,\xi)$.
Following \cite{LeM16}, the system is split into A,B,O components \eqref{eq:Acomponent}-\eqref{eq:Ocomponent}, where B represents a projected impulse defined by the loss gradient (restricted to the cotangent space), O represents a projected stochastic impulse, and A represents evolution along geodesics (i.e., for circle constraints, these are rotations on the circles).

As in the overdamped case, the equality $\nabla_{\theta^u}^Tg = 0_{m\times n^u}$ ensures that the unconstrained parameters and their momenta $(\theta^u,p^u)$ evolve following the A,B,O steps for unconstrained underdamped Langevin (see \cite{LMS16}).
As the B and O components only involve a variation in the momentum $p_t$ and because the constraint only involves $q_t$, they can be solved exactly for any constraint. 
The A component involves a variation of the configuration $q_t$ and thus cannot be solved exactly (in law) for any constraint.
However, as this part does not include any force evaluation (which would require back-propagation to compute the gradient), it can be approximated cheaply using a few steps of standard well-known schemes such as SHAKE or RATTLE (see Section \ref{sec:UL_orthogonalConstraint} for orthogonal constraints).
Furthermore, for simple constraints such as the circle constraint \eqref{eq:circleConstraint} the A component can be solved explicitly (see Section \ref{sec:discr_ULcircle}).

\def\confq{w}%
Let us present the details of the B and O steps.
For convenience, let us introduce the following notation for the variables involved in the constraint
$\confq=(\theta^c,\xi)\in\R^{n^c+n^\xi}$
and associated momentum 
$p^\confq=(p^c,p^\xi)\in\R^{n^c+n^\xi}$.
The projection onto the cotangent space \eqref{eq:OL_definitionPi} is then as 
\begin{equation}	\label{eq:expressionPiw}
\Pi(q) = I_d - 
\begin{pmatrix} 
0& 0 \\
0& \Pi_\confq(q)
\end{pmatrix},
\quad
\text{with }
\Pi_\confq = 
\begin{pmatrix} 
	g_{\theta^c}^T H^{-1} g_{\theta^c} 	&g_{\xi}^T H^{-1} g_{\theta^c} \\
	g_{\theta^c}^T H^{-1} g_{\xi}		&g_{\xi}^T H^{-1} g_{\xi} 
\end{pmatrix},
\end{equation}
where we have denoted the partial Jacobians by
$g_{\theta^c} = \nabla^T_{\theta^c}g\in\R^{m\times n^c}$,
$g_{\xi} = \nabla^T_{\xi}g\in\R^{m\times n^\xi}$
and the matrix $H = g_{\theta^c}g_{\theta^c}^T + g_{\xi}g_{\xi}^T \in\R^{m\times m}$.

{\bf B component}.
Given $q_0,p_0\in T^*\manifold$ and a time $t>0$, we have
\[
q_t = q_0,\qquad
p_t = p_0 - t \nabla_qV(q_0) - \nabla_q g(q_0)( \mu_t - \mu_0),
\]
where $\mu_t$ is such that $p_t\in T_{q_t}^*\manifold$ (i.e., it satisfies the constraint $0 = \nabla_q g(q_t)p_t$).
Note that as $q_0,p_0$ satisfy the constraints we have $\mu_0=0$.
Projecting onto the cotangent space $T_{q_t}^*\manifold = T_{q_0}^*\manifold$
and using $\Pi(q_0)\nabla_q g(q_0) = 0$ and $p_0=\Pi(q_0)p_0$, we obtain
\[
p_t = \Pi(q_t)p_t = \Pi(q_0)\big( p_0 - t \nabla_qV(q_0) - \nabla_qg(q_0) \mu_t \big)
= p_0 - t \Pi(q_0)\nabla_qV(q_0).
\]

The B step is thus obtained for a chosen stepsize $h>0$ as:
given $q_n = (\theta^u_n,\theta^c_n,\xi_n)\in\manifold$
and $p_n = (p^u_n,p^c_n,p^\xi_n)\in T_{q_n}^*\manifold$
\begin{equation}	\label{eq:UL_Bstep}
\text{(B, gen.)}\qquad
\begin{aligned}
&\theta^u_{n+1} =\theta^u_{n},
\qquad
\theta^c_{n+1} = \theta^c_{n},
\qquad
\xi_{n+1} = \xi_{n},
\\[5pt]
&
p^u_{n+1} = p^u_{n} - h \nabla_{\theta^u} \loss(\theta_n),
\qquad
\bar p^c_{n+1} = p^c_{n} - h \nabla_{\theta^c} \loss(\theta_n),
\qquad
\bar p^\xi_{n+1} = p^\xi_{n},
\\[5pt]
&
\begin{pmatrix} 
p^c_{n+1} \\  
p^\xi_{n+1}
\end{pmatrix}
= \Pi_{\confq}(\confq_n)
\begin{pmatrix} 
\bar p^c_{n+1}   \\
\bar p^\xi_{n+1}
\end{pmatrix}
\quad \text{where } \confq_n = \begin{pmatrix} \theta^c_{n}\\ \xi_{n} \end{pmatrix}.
\end{aligned}
\end{equation}

{\bf O component}.
Similarly as for the B part, the O part can be solved exactly in law for any constraint.
Given $q_0,p_0\in T^*\manifold$ and a time $t>0$, we have
\[
q_t = q_0,\qquad
p_t = p_0 - \gamma \int_0^t p_t\,\d t + \sqrt{2\gamma \tau}\int_0^t\,\d \mathcal{W}_t -\nabla_qg(q_0) \nu_t,
\]
where $\nu_t$ ensures that $p_t\in T_{q_t}^*\manifold$.
Projecting to the cotangent space $T_{q_t}^*\manifold = T_{q_0}^*\manifold$ as before, we obtain
\[
p_t = \Pi(q_t)p_t = p_0 - \gamma \int_0^t  \Pi(q_0)p_t\,\d t + \sqrt{2\gamma \tau}\Pi(q_0)\int_0^t\,\d \mathcal{W}_t.
\]
We thus recognize that $p_t$ is an Ornstein--Uhlenbeck process:
\[
p_t \stackrel{\text{law}}{=}   \Pi(q_0) \big( e^{-\gamma t} p_0 + \sqrt{\tau(1-e^{-2\gamma t})} R \big),
\qquad
\text{with }R \sim N(0,I_d),
\]
where the equality holds in law.

The O step is thus obtained for a chosen stepsize $h>0$ as:
given $q_n = (\theta^u_n,\theta^c_n,\xi_n)\in\manifold$
and $p_n = (p^u_n,p^c_n,p^\xi_n)\in T_{q_n}^*\manifold$
\begin{equation}	\label{eq:UL_Ostep}
\text{(O, gen.)}\qquad
\begin{aligned}
&\theta^u_{n+1} =\theta^u_{n},
\qquad
\theta^c_{n+1} = \theta^c_{n},
\qquad
\xi_{n+1} = \xi_{n},
\\[5pt]
&
p^u_{n+1} =  e^{-\gamma h} p^u_{n} + \sqrt{\tau(1-e^{-2\gamma h})}R^u,
\\&
\bar p^c_{n+1} =  e^{-\gamma h} p^c_{n} + \sqrt{\tau(1-e^{-2\gamma h})}R^c,
\\&
\bar p^\xi_{n+1} = e^{-\gamma h} p^c_{n} + \sqrt{\tau(1-e^{-2\gamma h})}R^\xi,
\\[5pt]
&
\begin{pmatrix} 
p^c_{n+1} \\  
p^\xi_{n+1}
\end{pmatrix}
= \Pi_{\confq}(\confq_n)
\begin{pmatrix} 
\bar p^c_{n+1}   \\
\bar p^\xi_{n+1}
\end{pmatrix}
\quad \text{where } \confq_n = \begin{pmatrix} \theta^c_{n}\\ \xi_{n} \end{pmatrix},
\end{aligned}
\end{equation}
and 
$R^u,R^c$, and $R^\xi$ are independent standard normal random variables.

\subsection{Circle constraint, overdamped Langevin (c-CoLA-od)}	\label{sec:discr_OLcircle}

We consider here the circle constraint \eqref{eq:circleConstraint}, for which the partial Jacobians are computed as
\begin{equation}	\label{eq:jacobianG_circle}
\nabla_q^T g = \big(\nabla_{\theta^u}^Tg, \nabla_{\theta^c}^Tg, \nabla_{\xi}^T g\big) \in\R^{m\times(n^u+n^c+m)},
\quad
\partial_{\theta^u_j} g_i = 0,
~~
\partial_{\theta^c_j} g_i = 2\theta^c_i\delta_{ij},
~~
\partial_{\xi_j} g_i = 2\xi_i\delta_{ij},
\end{equation}
where $\delta_{ij}$ is the Kronecker delta. 

For this constraint, the projection step in \eqref{eq:OLdiscretization1} can be computed explicitly.
Indeed $\lambda_{n}$ can be found by solving the $m$ quadratic equations
$0=g_i(\bar{q}_{n+1} -\nabla_q g(q_n)\lambda_n)$ $1\leq i\leq m$.
The (potential) two roots of each equation corresponds to the (potential) two projections of $\bar{q}_{n+1}$ onto the circle following the direction $\nabla g_i(q_n) = 2(\theta^c_{n,i},\xi_{n,i})$. When two roots are found, we may select the one closest to the point of origin $(\theta^c_{n,i},\xi_{n,i})$. 
However, if the point to project $(\bar\theta^c_{n+1,i},\bar\xi_{n+1,i})$ is too far away from the circle, this oblique projection may not be possible (i.e., the quadratic equation has no real root).

For the circle constraint, method \eqref{eq:OLdiscretization2} thus leads to a more robust projection process.
Indeed, as $\nabla g_i(q_{n+1}) = 2(\theta^c_{n+1,i},\xi_{n+1,i})$, the direction of the projection is now orthogonal to the circle.
To find an expression for the orthogonal projection $P$ of a point $(\bar\theta_1, \bar\xi_1)$ on the circle, it is easier to use a geometrical approach than to find the Lagrange multipliers:
\[
(\theta_1,\xi_1) = P (\bar\theta_1, \bar\xi_1) = 
\big(r_i\cos(\alpha), r_i\sin(\alpha)\big),
\quad\text{where }\alpha = \arctan\Big( \frac{\bar\xi_1}{\bar\theta_1}\Big).
\]

We obtain the following discretization of the overdamped Langevin with circle constraints.
We initialize the parameters of the neural network using standard PyTorch initialization \cite{Pytorch, Pytorchinit}, i.e., $\mathcal{U}(-1/\sqrt{N_{in}},1/\sqrt{N_{in}})$, where $N_{in}$ is the number of inputs to a layer. The auxiliary variables $\xi_i$ corresponding to the constrained parameters $\theta^c_i$ are initialized to obey the constraint $(\theta^c_i)^2 + \xi^2_i = r^2_i$. 
For a chosen stepsize $h>0$ and given a configuration $q_n = (\theta^u_n,\theta^c_n,\xi_n)\in\manifold$,
one step of the method is defined by $q_{n+1} = (\theta^u_{n+1},\theta^c_{n+1},\xi_{n+1})\in\manifold$ as
\begin{equation}	\label{eq:OL_circle_discr}
\begin{aligned}
&\theta^u_{n+1,i} 
	= \theta^u_{n,i} - h\partial_{\theta^u_i} \loss(\theta_n) + \sqrt{2\beta^{-1} h} R^u_i, \\
&\bar\theta^c_{n+1,i} 
	= \theta^c_{n,i} - h\partial_{\theta^c_i}\loss(\theta_n) + \sqrt{2\beta^{-1} h} R^c_i, \\
&\bar\xi_{n+1,i} 		
	= \xi_{n,i} + \sqrt{2\beta^{-1} h} R^\xi_i, \\
&\alpha_{n,i}
	= \arctan \left( \frac{\bar\xi_{n+1,i}}{\bar\theta^c_{n+1,i}} \right), \\
&\theta^c_{n+1,i} 
	 = r_i\cos(\alpha_{n,i}),\\
&\xi_{n+1,i} 		
	= r_i\sin(\alpha_{n,i}),
\end{aligned}
\end{equation}
where 
$R^u_i,R^c_i, R^\xi_i$ are independent standard normal random variables.


\subsection{Circle constraint, underdamped Langevin (c-CoLA-ud)}	\label{sec:discr_ULcircle}

We provide here the full discretization of the underdamped Langevin dynamics in the case of the circle constraint \eqref{eq:circleConstraint}.

\textbf{A component}.
For the circle constraint we can solve the A step explicitly.
First recall that as $\nabla^T_{\theta^u} g= 0$, the unconstrained parameters $\theta^u$ are obtained with a standard A step of the unconstrained underdamped Langevin.
Let us then focus on solving the constrained components: 
we denote $\confq=(\theta^c,\xi), p^\confq= (p^c,p^\xi)$.
Then for $1\leq i\leq m$ the A step in \eqref{eq:Acomponent} corresponds to the constrained ODEs
\begin{equation}	\label{eq:Acomponent_circle}
\begin{aligned}
&\dot{\confq}_i = p^\confq_i\\
&\dot{p}^\confq_i = -2 \lambda_i \confq_i\\
&|\theta^c_i|^2 + |\xi_i|^2 = r_i^2 ,\qquad
\theta^c_ip^c_i + \xi_i  p^\xi_i = 0.
\end{aligned}
\end{equation}
As these constrained ODEs are uncoupled, let us drop the specification of the index $i$.
By assumption, we are given initial conditions that satisfy the constraint $(\confq_0,p^\confq_0)\in T^*\manifold$.
Solving the second order ODE $\ddot{\confq} = -2\lambda \confq$, we find that any solution has the form
$\confq_t = R^{2\lambda}_t \confq_0$, where $R^{\omega}_t$ is a rotation matrix with angular speed $\omega$ given with its time derivative as
\[
R^\omega_t = \begin{pmatrix} \cos(\omega t) &\sin(\omega t) \\ -\sin(\omega t) & \cos(\omega t) \end{pmatrix},
\qquad
\dot{R}^\omega_t = \omega \begin{pmatrix} -\sin(\omega t) &\cos(\omega t) \\ -\cos(\omega t) & -\sin(\omega t) \end{pmatrix}.
\]
Computing the momentum $p^\confq_t = \dot{\confq}_t = \dot{R}^\omega_t \confq_0$,
and using the properties of $R^\omega_t$
we verify that $\confq_t,p^\confq_t$ satisfy the constraints in \eqref{eq:Acomponent_circle}
($\|.\|$ denotes the Euclidean norm in $\R^2$ and $\cdot$ the dot product):
\[
\|\confq_t\|^2 = \|R^\omega_t \confq_0\|^2 = \|\confq_0\|^2 = r^2,
\qquad
\confq_t\cdot p^\confq_t = \confq_0^T(R^\omega_t)^T \dot{R}^\omega_t\confq_0 = 0. 
\]
We still have to find the angular speed $\omega=2\lambda$ such that the momentum $p^\confq_t$ is consistent with its initial value $p^\confq_0$ (we denote $\confq_0=(\theta^c_0,\xi_0)$ and $p^\confq_0= (p^c_0,p^\xi_0))$:
\[
p^\confq_0 = \dot{R}^\omega_0 u_0 
\quad\Leftrightarrow\quad
p^c_0 = \omega \xi_0
~\text{ and }~
p^\xi_0 = -\omega \theta^c_0.
\]
We thus find that 
\[
\xi_0 p^c_0 -\theta^c_0 p^\xi_0 = \omega \big(|\xi_0|^2+|\theta^c_0|^2\big) = \omega r^2
\quad\Leftrightarrow\quad
\omega = \frac{1}{r^2} \big(  \xi_0 p^c_0 - \theta^c_0 p^\xi_0 \big).
\]
We have thus found an explicit expression for the solution of the A component for circle constraints \eqref{eq:Acomponent_circle}.

To complete the B and O steps given in \eqref{eq:UL_Bstep} and \eqref{eq:UL_Ostep}, we need an explicit expression for the projection $\Pi_\confq$ in \eqref{eq:expressionPiw} (using \eqref{eq:jacobianG_circle}, recall that $m=n^c=n^\xi$):
\[
\Pi_\confq(w) = 
\begin{pmatrix}
I_m - D^{11} 			& - D^{12} \\
- D^{12}					&I_m - D_{22}
\end{pmatrix},
\]
where $D^{kl}\in\R^{m\times m}$ are the diagonal matrices defined as
\[
D^{11}_{ii} = \frac{|\theta^c_i|^2}{ |\theta^c_i|^2 + |\xi_i|^2 },
\quad
D^{12}_{ii} = \frac{\theta^c_i\xi_i}{ |\theta^c_i|^2 + |\xi_i|^2 },
\quad
D^{22}_{ii} = \frac{|\xi_i|^2}{ |\theta^c_i|^2 + |\xi_i|^2 }.
\]
Assuming that $\confq=(\theta^c,\xi)$ satisfies the constraint,
the projection of $(\bar{p}^c,\bar{p}^\xi)$ is thus computed as
\[
\begin{pmatrix}{p}^c \\{p}^\xi\end{pmatrix} 
	= \Pi_\confq (w)
		\begin{pmatrix}\bar{p}^c \\\bar{p}^\xi\end{pmatrix}
,
\quad\text{where} \qquad
\begin{aligned}
{p}^c_i		
	&= \bar{p}^c_i   -\frac{\theta^c_i}{r_i^2}\big( \theta^c_i \bar{p}^c_i +\xi_i \bar{p}^\xi_i\big)
		\quad 1\leq i\leq m,\\
{p}^\xi_i 
	&= \bar{p}^\xi_i -\frac{\xi_i}{r_i^2}\big( \theta^c_i \bar{p}^c_i +\xi_i \bar{p}^\xi_i\big)
		\quad 1\leq i\leq m.
\end{aligned}
\]
Note that in the B step \eqref{eq:UL_Bstep}, the above expressions can be simplified 
by combining the simple definition of $(\bar{p}^c_n,\bar{p}^\xi_n)$ with the constraint 
\[
0 = \big(\nabla^Tg(q)p\big)_i = 2 \big(\theta^c_i p^c_i + \xi_i p^\xi_i \big).
\]

We provide below the explicit updates for the A, B and O components for circle constraints. We initialize the parameters of the net using standard PyTorch initialization \cite{Pytorch, Pytorchinit}. The auxiliary variables $\xi$ corresponding to the constrained parameters $\theta^c$ are initialized to obey the constraint $(\theta^c)^2 + \xi^2 = r^2$, so that $q_0 = (\theta^u_0,\theta^c_0,\xi_0)\in\manifold$. The momenta, $p^u, p^c$, and $p^{\xi}$, are generated in the same manner as for standard SGD with momentum in PyTorch, i.e., as equal to the initial gradients. Subsequently, the momenta belonging to the constrained variables $p^c$ and to the auxiliary variables $p^{\xi}$ are projected using $\Pi_\confq$, so that $p_0 = (p^u_0,p^c_0,p^\xi_0)\in T_{q_0}^*\manifold$.
For a stepsize $h>0$ we obtain
\begin{equation*}
\text{(A step, circle)}\qquad
\left\{
\begin{aligned}
&\theta^u_{n+1,i} 
= \theta^u_{n,i} + h p^u_{n,i},
\quad
\\
&\omega_i = \frac{1}{r_i^2} \big(  \xi_{n,i} p^c_{n,i} - \theta^c_{n,i} p^\xi_{n,i} \big),
\\
&\theta^c_{n+1,i} 
= \cos(\omega_i h)\theta^c_{n,i}  + \sin(\omega_i h)\xi_{n,i}  , 
\\
&\xi_{n+1,i} 		
= -\sin(\omega_i h)\theta^c_{n,i}  + \cos(\omega_i h)\xi_{n,i} ,
\\[5pt]
&p^u_{n+1,i} = p^u_{n,i},
\\
&p^c_{n+1,i} = \omega_i\big(-\sin(\omega_i h)\theta^c_{n,i}  + \cos(\omega_i h)\xi_{n,i}\big),
\\
&p^\xi_{n+1,i} = -\omega_i\big(\cos(\omega_i h)\theta^c_{n,i}  +\sin(\omega_i h)\xi_{n,i}\big),
\end{aligned}
\right.
\end{equation*}
\begin{equation*}
\text{(B step, circle)}\qquad
\left\{
\begin{aligned}
&\theta^u_{n+1} =\theta^u_{n},
\qquad
\theta^c_{n+1} = \theta^c_{n},
\qquad
\xi_{n+1} = \xi_{n},
\\[5pt]
&
p^u_{n+1} = p^u_{n} - h \nabla_{\theta^u} \loss(\theta_n),
\\&
\bar p^c_{n+1,i} = p^c_{n,i} - h\Big( 1 - \frac{1}{r_i^2}|\theta^c_{n,i}|^2 \Big) \partial_{\theta^c_i} \loss(\theta_n),
\\&
\bar p^\xi_{n+1,i} = p^\xi_{n,i} + h\frac{1}{r_i^2}\theta^c_{n,i}\xi_{n,i} \partial_{\theta^c_i} \loss(\theta_n),
\end{aligned}
\right.
\end{equation*}

\begin{equation*}	
\text{(\text{O step, circle})}\qquad
\left\{
\begin{aligned}
&\theta^u_{n+1} =\theta^u_{n},
\qquad
\theta^c_{n+1} = \theta^c_{n},
\qquad
\xi_{n+1} = \xi_{n},
\\[5pt]
&
p^u_{n+1} =  e^{-\gamma h} p^u_{n} + \sqrt{\beta^{-1}(1-e^{-2\gamma h})}R^u,
\\&
\bar p^c_{n+1} =  e^{-\gamma h} p^c_{n} + \sqrt{\beta^{-1}(1-e^{-2\gamma h})}R^c,
\\&
\bar p^\xi_{n+1} = e^{-\gamma h} p^c_{n} + \sqrt{\beta^{-1}(1-e^{-2\gamma h})}R^\xi,
\\
&
p^c_{n+1,i}	
= 	\Big(1 - \frac{1}{r_i^2} |\theta^c_{n,i}|^2\Big)   \bar  p^c_{n+1,i}  
-  \frac{1}{r_i^2} \theta^c_{n,i}\xi_{n,i} \bar  p^\xi_{n+1,i},
\\
&
p^\xi_{n+1,i} 
= 	-\frac{1}{r_i^2} \theta^c_{n,i}\xi_{n,i} \bar  p^c_{n+1,i}  
+ \Big(1 - \frac{1}{r_i^2} |\xi_{n,i}|^2\Big) \bar  p^\xi_{n+1,i},
\end{aligned}
\right.
\end{equation*}
where $R^u,R^c$, and $R^\xi$ are vectors of independent standard normal random variables.

\subsection{Orthogonality constraint, overdamped Langevin dynamics (o-CoLA-od)}	\label{sec:OL_orthogonalConstraint}

We present here a particular discretization of the constrained overdamped Langevin dynamics \eqref{eq:constrainedOverdampedLangevin} for the orthogonality constraint \eqref{eq:orthogonalConstraint}.

For notational convenience, we present the updates for the weight matrix $W^\ell$ of a given layer $\ell$.
The updates for the biases are standard Euler--Maruyama steps  such as given for $\theta^u$ in \eqref{eq:OL_circle_discr}.
 
Referring to \eqref{eq:orthogonalConstraint}, we denote
\begin{equation}
\begin{array}{llll}
Q = W^\ell,& r=n^{\ell}, &s = n^{\ell-1}	&\text{if }n^{\ell-1}\leq n^{\ell},\\
Q=(W^\ell)^T,& r=n^{\ell-1}, &s = n^{\ell}	&\text{otherwise}.
\end{array}
\end{equation}
so that $Q\in\R^{r\times s}$.
With this notation, the constraint \eqref{eq:orthogonalConstraint} is $g(Q)=0$ where
\begin{equation}	\label{eq:orthogonalConstraint_g}
g : \R^{r\times s} \to \R^{s\times s}, \qquad g(Q) = Q^TQ - I_s.
\end{equation}
Recall that due to symmetry, the matrix equality $g(Q)=0_s$ corresponds to $s(s+1)/2$ constraints.
We compute the partial derivative
\begin{equation}	\label{eq:partialDerivativeOGconstraint}
\partial_{Q_{kl}} g_{ij}(Q) = \delta_{li} Q_{kj} + \delta_{lj} Q_{ki}
\qquad
1\leq i,j,k\leq s, ~1\leq l\leq r.
\end{equation}
In particular, if $\Lambda$ is an $s\times s$ symmetric matrix, we verify that 
\[
\sum_{i,j=1}^s \partial_{Q_{kl}} g_{ij}(Q) \Lambda_{ij}  = 2\big( Q\Lambda\big)_{kl}.
\]
We thus obtain the natural matrix form of the constrained dynamics \eqref{eq:constrainedOverdampedLangevin}:
$Q_t:(0,\infty)\to \R^{r\times s}$ solves
\begin{equation}	\label{eq:continuousSDE_orthogonalConstraint}	
\begin{aligned}
&\d Q_t = -\nabla_Q V(Q_t) \,\d t + \sqrt{2\beta^{-1}}\,\d \mathcal{W}_t  - Q_t \,\d \Lambda_t ,\\
&g(Q_t) = 0,
\end{aligned}
\end{equation}
where
$\big(\nabla_Q V\big)_{ij} = \partial_{Q_{ij}} V = \partial_{W^\ell_{ij}} \loss$ (or $\partial_{W^\ell_{ji}} \loss$)
and $\mathcal{W}_t$ is a Wiener process in $\R^{r\times s}$.
Furthermore the process $\Lambda_t$ has values in the $s\times s$ symmetric matrices and is the Lagrange multiplier corresponding to the $s(s+1)/2$ constraints.

Applying discretization scheme \eqref{eq:OLdiscretization1} to \eqref{eq:continuousSDE_orthogonalConstraint}, we obtain the iteration step
$Q_n\in\manifold \mapsto Q_{n+1}\in\manifold$ given by
\begin{equation}	\label{eq:OLdiscr_orthogonalConstraint}
\begin{aligned}
&\bar Q_{n+1} = Q_n - h\nabla_Q V(Q) + \sqrt{2\beta^{-1}h} R_n,
\qquad
Q_{n+1} = \bar{Q}_{n+1} - Q_n\Lambda_n,\\
&\text{where $\Lambda_n$ is a symmetric $s\times s$ matrix s.t. $g(Q_{n+1})=0$},
\end{aligned}
\end{equation}
and $R_n\in\R^{r\times s}$ is a matrix of independent standard normal random variables.

Note that the projection step in \eqref{eq:OLdiscr_orthogonalConstraint} requires to solve a non-linear system.
Following a similar technique as described in \cite[Chap. 8]{LeR04}, we derive a quasi-Newton scheme for that task. 
Using the fact that $Q_n$ satisfies the constraint we verify that
\[
\bar{Q}_ {n+1}^T Q_n = I_s - h\nabla_Q V(Q_n)^TQ_n + \sqrt{2\beta^{-1}h} R_n^TQ_n.
\]
The constraint $g(Q_{n+1})=0$ thus reads
\begin{equation}	\label{eq:estimateQuasiNewton}
0 	
	= \big(\bar{Q}_{n+1} - Q_n\Lambda_n\big)^T \big(\bar{Q}_{n+1} - Q_n\Lambda_n\big)-I_s
	= \big(\bar{Q}_{n+1}^T\bar{Q}_{n+1} - I_s\big) - 2\Lambda_n + \mathcal{O}(\sqrt{h}),
\end{equation}
where $\mathcal{O}(\sqrt{h})$ denotes a matrix whose 2-norm has order $\sqrt{h}$.
Solving for $\Lambda_n$, we find
\[
\Lambda_n = \frac{1}{2} \big( \bar{Q}_{n+1}^T\bar{Q}_{n+1} - I_s \big)  + \mathcal{O}(\sqrt{h}).
\]
Neglecting the terms of order $\sqrt{h}$ and higher, we obtain the following quasi-Newton scheme:
setting $Q^{(0)}=\bar{Q}_{n+1}$, repeat the iteration
\begin{equation}	\label{eq:quasiNewton_orthogonalConstraint}
Q^{(k+1)} = Q^{(k)} - Q_n\Lambda^{(k)}, 
\quad\text{where } \Lambda^{(k)} = \frac{1}{2} \big( (Q^{(k)})^TQ^{(k)} - I_s \big),
\end{equation}
until the process reaches convergence and set $Q_{n+1} = Q^{(k+1)}$.
To assess whether convergence has been reached, a tolerance on the $2$-norm of $\Lambda^{(k)}$ can be assigned:
$\|\Lambda^{(k)}\| \leq \mathrm{TOL}$.
However in practice, to ensure that the process ends and to avoid undesirable overhead we typically prefer to either combine this stopping criterion with a limit for the number $K$ of iterations, or use a fixed number of iterations $K$.
Note that estimate \eqref{eq:estimateQuasiNewton} ensures that a small number of iterations $K$ is sufficient for the constraint to be satisfied up to a small error.

The initialization for the constrained weights is performed following \cite{Saxe2013}, which is an built-in option in PyTorch. Other parameters are initialized using the standard PyTorch initialization \cite{Pytorch, Pytorchinit} unless otherwise indicated. Constraints are applied layer-wise, where for convolutional layers with weight tensors of the size $n_l \times n_{l-1} \times n_h \times n_w$ (where $n_h$ and $n_w$ are the height and width of the kernel) the weight matrices are reshaped as $n_l \times n_{l-1} n_h n_w$. For CNNs these reshaped matrices are typically rectangular. If they are thin, but long (i.e., $n_l > n_{l-1} n_h n_w$) we apply the constraint $W^T W = I$, but if they have more columns than rows we apply the constraint $W W^T = I$. 

\subsection{Orthogonality constraint, underdamped Langevin (o-CoLA-ud)}	\label{sec:UL_orthogonalConstraint}

To discretize the underdamped Langevin constrained dynamics, we need the orthogonal projection $\Pi$ onto the cotangent space $T_Q^*\manifold$. As the constraint \eqref{eq:orthogonalConstraint_g} is given in a matrix form, using the formula \eqref{eq:OL_definitionPi} is not very convenient so we will rather derive $\Pi$ from its projection property.

Using \eqref{eq:partialDerivativeOGconstraint}, we find that for $1\leq i,j\leq s$
\[
0 = \sum_{k=1}^s\sum_{l=1}^r \partial_{Q_{kl}} g_{ij}(Q) P_{kl} = (P^TQ + Q^TP)_{ij},
\]
which leads to the following convenient expression for the cotangent space 
\[
T^*_Q\manifold = \big\{  P \in\R^{r\times s} \mid P^TQ + Q^TP = 0_s \big\}.
\]
Now, given $\bar P\in\R^{r\times s}$ we want to find a symmetric $s \times s$ matrix $\Lambda$ such that 
$P = \bar P - Q\Lambda$ belongs to $T^*_Q\manifold$, i.e.,
\[
0_s = P^TQ -Q^TP =  \bar P^TQ + Q^T\bar P - \Lambda Q^TQ - Q^TQ \Lambda.
\]
This equation is easily solved for $Q\in\manifold$ and we find 
$\Lambda = \frac{1}{2} (\bar P^TQ + Q^T\bar P)$. We obtain the following expression for the projection onto the cotangent space:
\[
\Pi_Q : \R^{r\times s} \to \R^{r\times s} ,\quad
\bar P \mapsto \Pi_Q\bar P = \bar P - \frac{1}{2} Q (\bar P^TQ + Q^T\bar P).
\]
We then verify that $\Pi_Q$ is indeed a projection onto the cotangent space $T_Q^*\manifold$
(i.e., $\Pi_Q\bar P\in T_Q^*\manifold$ $\forall\bar P\in\R^{r\times s}$ and $\Pi_Q^2 = \Pi_Q$) 
and that this projection is orthogonal with respect to the Frobenius inner product on $\R^{r\times s}$ 
(i.e., $\langle \bar P-\Pi_Q\bar P, P \rangle = 0$, where $\langle A,B\rangle = \mathrm{tr}(A^TB)$).
\ \\ \ \\ 
\textbf{A component}.
For the orthogonal constraint, the A component in \eqref{eq:Acomponent} can only be solved approximately.
A simple yet efficient discretization of A is the RATTLE scheme (see e.g. \cite[Chap. 8]{LeR04}):
\begin{equation}	\label{eq:ogConstraint_A_RATTLE0}
\begin{aligned}
Q_{n+1} &= Q_n  + hP_{n+1/2},\\
P_{n+1/2} &= P_n - Q_n\Lambda_{n+1/2}
	\quad \text{where $\Lambda_{n+1/2}$ is s.t. }Q_{n+1}^TQ_{n+1} = I_s,\\
P_{n+1} &= P_{n+1/2} - Q_{n+1} \Lambda_{n+1} 
	\quad \text{where $\Lambda_{n+1}$ is s.t. }Q_{n+1}^TP_{n+1} + P_{n+1}^T Q_{n+1} = 0_s.
\end{aligned}
\end{equation}
Denoting $\bar\Lambda_{n+1/2} = h\Lambda_{n+1/2}$, $\bar P_{n+1}=P_{n+1/2}$ and using the projection operator $\Pi_Q$, \eqref{eq:ogConstraint_A_RATTLE0} can be rewritten as  
\begin{equation}
\begin{aligned}
\bar Q_{n+1} &= Q_n  + hP_{n},\\
Q_{n+1} &= \bar Q_{n+1} - Q_n\bar\Lambda_{n+1/2}
\quad \text{where $\bar\Lambda_{n+1/2}$ is s.t. }Q_{n+1}^TQ_{n+1} = I_s 
\quad \text{(use \eqref{eq:quasiNewton_orthogonalConstraint})},\\
\bar P_{n+1} &=P_n - \frac{1}{h} Q_n\bar\Lambda_{n+1/2},\qquad
P_{n+1} = \Pi_{Q_{n+1}} \bar P_{n+1}.
\end{aligned}
\end{equation}
As in the overdamped case, we may now use the quasi-Newton scheme \eqref{eq:quasiNewton_orthogonalConstraint} for the projection step (to approximate $\bar\Lambda_{n+1/2}$).
Using $K$ iterations of the quasi-Newton scheme \eqref{eq:quasiNewton_orthogonalConstraint} (i.e., $Q_{n+1}=Q^{(K)}$),
we verify that $-Q_n\bar\Lambda_{n+1/2}$ satisfies
\[
- Q_n\bar\Lambda_{n+1/2} 
	= \sum_{k=0}^{K-1} Q_n\Lambda^{(k)} 
	= \sum_{k=0}^{K-1} Q^{(k+1)}- Q^{(k)}
	= Q^{(K)}-Q^{(0)}
	= Q_{n+1} - \bar Q_{n+1},
\]
so that $\bar P_{n+1} =P_n + \frac{1}{h}(Q_{n+1} - \bar Q_{n+1})$.

We obtain the following full discretization of the underdamped Langevin dynamics with orthogonality constraint.
The initialization for the constrained weights is performed following \cite{Saxe2013}. Corresponding momenta are initialized as the initial gradients (equivalently to standard PyTorch initialization) and subsequently projected using $P_0 = \bar{P}_0 - \frac{1}{2}Q_0(\bar{P}^T_0 Q_0 + Q_0^T \bar{P}_0)$. The A,B,O steps are then given as:

\begin{equation*}	
\text{(\text{A, OG})}\qquad
\left\{
\begin{aligned}
&\bar Q_{n+1} = Q_n  + hP_{n},\quad Q^{(0)} = \bar Q_{n+1}, \\
&\text{for $k = 0:K-1$} \qquad Q^{(k+1)} = Q^{(k)} - Q_n\Lambda^{(k)}, 
	\quad\text{where } \Lambda^{(k)} = \frac{1}{2} \Big( \big(Q^{(k)}\big)^T Q^{(k)} - I_s \Big), \\
&Q_{n+1} = Q^{(K)},\\
&\bar P_{n+1} =P_n + \frac{1}{h}\big(Q_{n+1}-\bar Q_{n+1}\big), \\
&P_{n+1} = \Pi_{Q_{n+1}} \bar P_{n+1}
	= \bar{P}_{n+1}  - \frac{1}{2}Q_{n+1} \Big(\bar{P}^T_{n+1} Q_{n+1} + \big(Q_{n+1}\big)^T \bar{P}_{n+1}\Big).
\end{aligned}
\right.
\end{equation*}
\begin{equation*}
\text{(B, OG)}\qquad
\left\{
\begin{aligned}
&Q_{n+1} =Q_{n},
\\
&\bar P_{n+1} = P_n - h\nabla_Q V(Q_n),\quad
\\&P_{n+1} = \Pi_{Q_n}P_{n+1} 
	= \bar{P}_{n+1}  - \frac{1}{2}Q_{n} \Big(\bar{P}^T_{n+1} Q_{n} + \big(Q_{n}\big)^T \bar{P}_{n+1}\Big),
\end{aligned}
\right.
\end{equation*}
\begin{equation*}	
\text{(\text{O, OG})}\qquad
\left\{
\begin{aligned}
&Q_{n+1} =Q_{n},
\\
&\bar P_{n+1} = e^{-\gamma h} P_{n} + \sqrt{\beta^{-1}(1-e^{-2\gamma h})}R_n,
\\&P_{n+1} = \Pi_{Q_n} \bar P_{n+1}
 	=  \bar{P}_{n+1}  - \frac{1}{2}Q_{n} \Big(\bar{P}^T_{n+1} Q_{n} + \big(Q_{n}\big)^T \bar{P}_{n+1}\Big),
\end{aligned}
\right.
\end{equation*}
where $R_n$ is a matrix of independent standard normal random variables.

\section{Additional Numerical Details and Results}\label{sec:NumericsAppx}
We compare our constrained methods with PyTorch's \cite{Pytorch} SGD with momentum optimiser. Unless otherwise indicated, we use for SGD $h = 0.1$ and $mom = 0$ (to compare with our constrained overdamped Langevin method) or $mom = 0.9$ (to compare with our constrained underdamped Langevin method). We use standard PyTorch initialization for all unconstrained parameters \cite{Pytorchinit, Pytorch}. 

\subsection{Orthogonality Constraints}
\begin{figure}[h]
    \centering
    \includegraphics[scale=0.3]{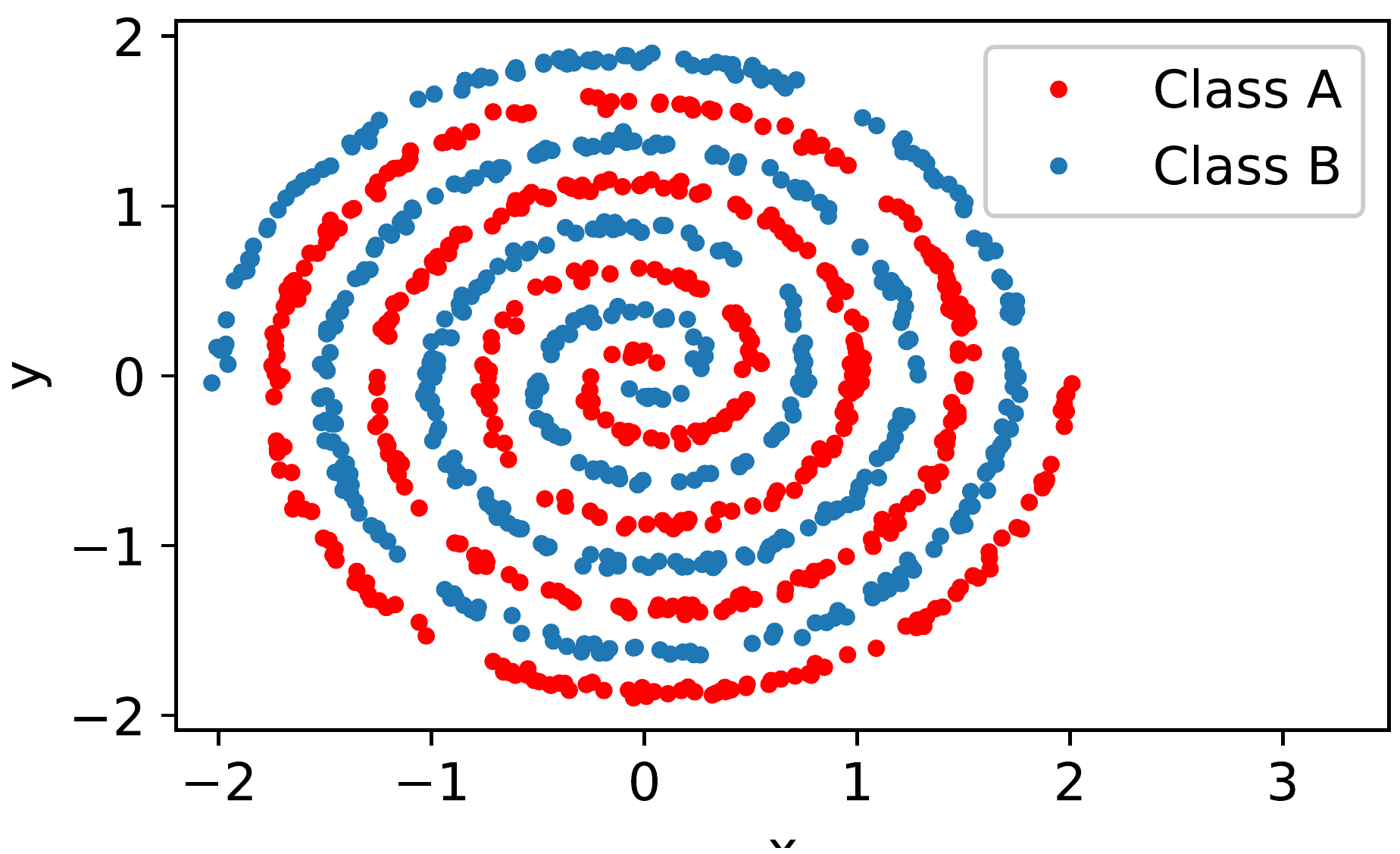}\caption{4-turn spiral data set.} \label{Spiraldata}
\end{figure}

For our experiments on the spiral data set (see Fig. C\ref{Spiraldata}) we use multi-layer perceptrons with ReLU activation and binary cross entropy loss. In our experiments we vary the number of 100-node hidden layers of the multi-layer perceptrons. To compare the performance of our o-CoLA-od constrained method with standard SGD we set the temperature $\tau = 0$ to generate Figure \ref{MLP_OCSGD}. 
We used a small temperature perturbation $\tau$ = 1e-6 to generate Figure \ref{Spiral_withT}. The size of the temperature parameter was chosen to approximately match observed fluctuations in the loss function. A more precise parameterization is left for a subsequent work.

\ \\ \ \\ 
A plot of the planar spiral data set binary classification problem as used to produce Figure \ref{MLP_OCSGD} and Figure \ref{Spiral_withT} is provided in Figure C\ref{Spiraldata}. The first class of the data set is generated using 
\begin{align}
 x& = 2 \sqrt{t} \cos(8 \sqrt{t} \pi) +0.02\mathcal{N}(0,1), \nonumber \\
 y& = 2 \sqrt{t} \sin(8 \sqrt{t} \pi) +0.02\mathcal{N}(0,1), \label{spiraleqn}
\end{align}
where $t$ is drawn repeatedly from the uniform distribution $\mathcal{U}(0,1)$ to generate data points. The other class of this dataset is obtained by shifting the argument of the trigonometric functions by $\pi$. For our experiments we used 500 training data points, 1000 test data points and 5\% subsampling. 
 \\
\makeatletter
\renewcommand{\fnum@figure}{\figurename~C\thefigure}
\makeatother

\begin{figure}[!h]
     \centering
       \includegraphics[width=0.8\linewidth]{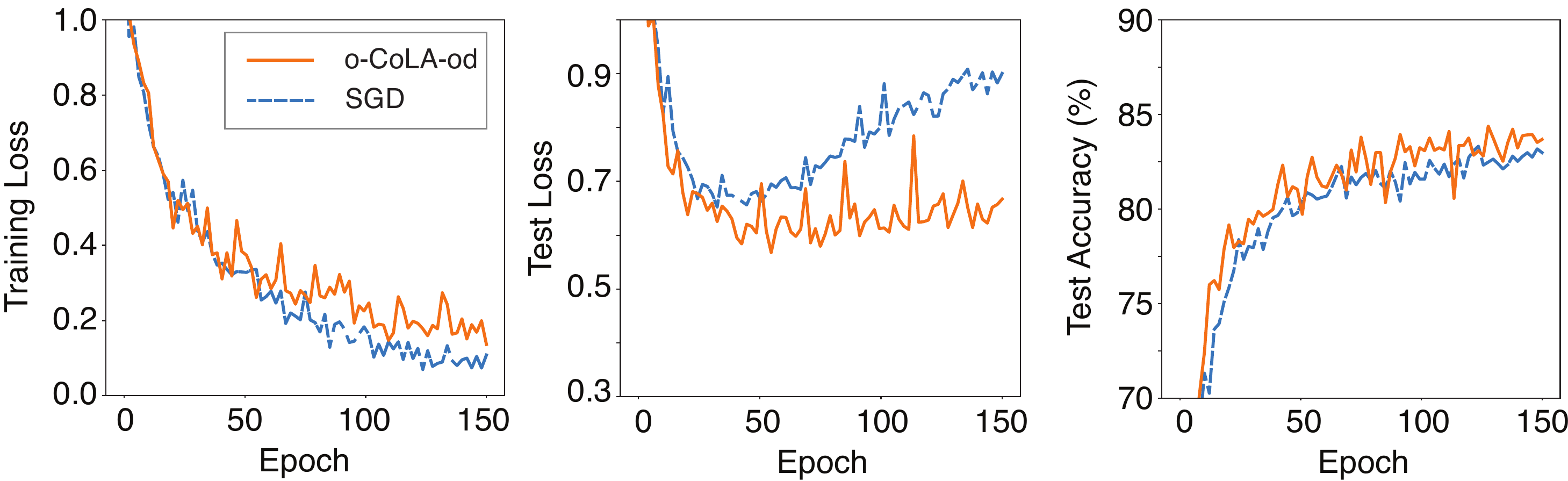}
       \vspace{-1mm}
     \caption{Training loss (left), test loss (middle) and test accuracy (right) of a ResNet-34 architecture trained using SGD vs. o-CoLA-od on CIFAR-10 data, $h = 0.1$ (averaged over 5 runs). The orthogonality constraint provides modestly higher test accuracy and inhibits overfitting.\label{CSGD_Resnet}}
\end{figure}

We also applied our orthogonality-constrained methods to the ResNet-34 architecture on CIFAR-10 image classification data \cite{cifar10}. In this setting, running SGD with orthogonal initialization worsened the generalization performance of the resulting net and hence the standard PyTorch initialization was used for SGD. We train for 150 epochs and use a batchsize of 128. In Figure C\ref{CSGD_Resnet} we compare the overdamped variant o-CoLA-od (with $\tau = 0$) to its unconstrained counterpart. We observe that constraining orthogonality gives lower test loss throughout training.

\subsection{Circle constraints}\label{tableFashionMNIST}
For our Fashion-MNIST \cite{FashionMNIST} example we reduce the number of training data samples to 10,000 and we increase the number of test data samples to 60,000. We use a 1000-node single hidden layer perceptron (SHLP) with ReLU activation, cross entropy loss and batchsize 128. Our main result with our circle constrained approach is presented in Figure \ref{CNet_FashionMNIST}. In this section we present extensive hyperparameter tests for the test accuracy and test loss obtained after 400 epochs (averaged over 5 runs) using SGD-m with and without weight decay (WD). \\
\begin{tabular}{c|c|c|c|c|c}
  \multicolumn{2}{c|}{} & \multicolumn{2}{c|}{no WD} & \multicolumn{2}{c}{with WD} \\ 
    \multicolumn{2}{c|}{SGD with mom} & Test Acc. & Test Loss & Test acc. & Test Loss  \\ \hline
    h = 0.2 & mom = 0.8  & 87.18\% & 1.06 & 84.05\% & 0.696 \\
    & mom = 0.7  & 87.38\% & 0.890 & 87.0\% & 0.547 \\
     \hline
    h = 0.1  & mom = 0.9 & 86.97\% & 1.133 & 85.35\% & 0.634 \\ 
    & mom = 0.8 & 87.39\% & 0.824 & 87.47\% & 0.531 \\
             & mom = 0.7 & 87.39\% & 0.750 & 87.25\% & 0.517 \\ \hline
    h = 0.05 & mom = 0.95 & 86.67\% & 1.226 & 85.63\% & 0.623\\
    & mom = 0.9 & 87.33\% & 0.837 & 86.24\% & 0.569 \\
             & mom = 0.8 & 87.27\% & 0.719 & 87.33\% & 0.511\\
\end{tabular}
\ \\ \ \\
The results presented in the two right-hand columns are all obtained with weight decay set to \\ 1e-4. We found this value to give the best results for SGD-m during a hyperparameter search. \ \\ \ \\ 
In comparison our circle constrained net reaches test accuracy \textbf{87.61\%}, with test loss \textbf{0.386} without using weight decay (see Figure \ref{CNet_FashionMNIST}). Hence it outperforms standard SGD with momentum both with and without weight decay. 

\subsection{NLP}
We evaluate the performance of a small transformer model \cite{Transformer} on the Penn Treebank data set \cite{PennTreebank} and Wikitext-2 data set \cite{Wikitext2}. The transformer has 2 encoder layers. Each encoder layer consists of self- attention with 2 heads and a feedforward network with 200 nodes followed by layer norms. We use batch size 1024 for the Penn Treebank data set and batchsize 128 for the Wikitext-2 dataset. We present the lowest validation loss obtained in 200 epochs by SGD-m and our circle constrained method c-CoLA-ud in Table \ref{NLPexp} of the main paper.

\end{document}